\documentclass[10pt,twocolumn,letterpaper]{article}

\usepackage{iccv}
\usepackage{times}
\usepackage{epsfig}
\usepackage{graphicx}
\usepackage{amsmath}
\usepackage{amssymb}
\usepackage{lipsum}
\usepackage{subcaption}
\usepackage{xcolor}
\usepackage{multicol}
\usepackage{multirow}
\usepackage{tabularx}
\usepackage[pangram]{blindtext}
\usepackage{enumerate}
\usepackage{cite}
\usepackage{booktabs}
\usepackage{colortbl}
\usepackage{caption}
\definecolor{darkgreen}{rgb}{0, 0.5, 0}
\definecolor{myblue}{RGB}{47, 114, 193}
\definecolor{brickred}{rgb}{0.8, 0.25, 0.33}
\definecolor{brandeisblue}{rgb}{0.0, 0.44, 1.0}
\definecolor{blueish}{rgb}{0.0, 0.3, .6}
\definecolor{pink}{rgb}{1, 0, 1}

\usepackage[pagebackref=true,breaklinks=true,colorlinks,citecolor=blueish,linkcolor=brickred,bookmarks=false]{hyperref}

\usepackage[capitalize]{cleveref}
\crefname{section}{Sec.}{Secs.}
\Crefname{section}{Section}{Sections}
\Crefname{table}{Table}{Tables}
\crefname{table}{Tab.}{Tabs.}

\iccvfinalcopy % *** Uncomment this line for the final submission

\def\iccvPaperID{****} % *** Enter the ICCV Paper ID here

\ificcvfinal\fi

\begin{document}

%%%%%%%%% TITLE
% \title{\LaTeX\ Author Guidelines for ICCV Proceedings}
\title{MULLER: Multilayer Laplacian Resizer for Vision}

\author{
Zhengzhong Tu, Peyman Milanfar, Hossein Talebi \\
Google Research\\
}

\maketitle
\ificcvfinal\thispagestyle{empty}\fi

%%%%%%%%% ABSTRACT
\begin{abstract}
Image resizing operation is a fundamental preprocessing module in modern computer vision. Throughout the deep learning revolution, researchers have overlooked the potential of alternative resizing methods beyond the commonly used resizers that are readily available, such as nearest-neighbors, bilinear, and bicubic. The key question of our interest is whether the front-end resizer affects the performance of deep vision models? In this paper, we present an extremely lightweight multilayer Laplacian resizer with only a handful of trainable parameters, dubbed MULLER resizer. MULLER has a bandpass nature in that it learns to boost details in certain frequency subbands that benefit the downstream recognition models. We show that MULLER can be easily plugged into various training pipelines, and it effectively boosts the performance of the underlying vision task with little to no extra cost. Specifically, we select a state-of-the-art vision Transformer, MaxViT~\cite{tu2022maxvit}, as the baseline, and show that, if trained with MULLER, MaxViT gains up to 0.6\% top-1 accuracy, and meanwhile enjoys 36\% inference cost saving to achieve similar top-1 accuracy on ImageNet-1k, as compared to the standard training scheme. Notably, MULLER's performance also scales with model size and training data size such as ImageNet-21k and JFT, and it is widely applicable to multiple vision tasks, including image classification, object detection and segmentation, as well as image quality assessment.

\end{abstract}

%%%%%%%%% BODY TEXT
\section{Introduction}

Most computer vision problems such as image classification, object detection, video recognition, and image/video generation have seen groundbreaking advancement by deep neural networks-based models that are trained on web-scale, human-curated datasets~\cite{krizhevsky2012imagenet,he2016deep,szegedy2015going,krizhevsky2012imagenet,vaswani2017attention,szegedy2016rethinking,huang2017densely,liu2021swin,xu2022v2x,sandler2018mobilenetv2,tan2019efficientnet}.
In any of the underlining training infrastructures like Tensorflow~\cite{abadi2016tensorflow} and PyTorch~\cite{paszke2019pytorch}, image resizing is an essential preprocessing step which enables efficient gradient-based training of networks with millions of trainable parameters.
Moreover, the factor of image size can sometimes significantly impact the performance of various tasks, particularly those requiring high-resolution prediction.
Although neural architectures have been revolutionized by CNNs and Transformers, surprisingly limited attention has been paid to the role of image resizing operations.

\begin{figure}[!t]
\centering

\begin{tabular}{c}
\includegraphics[width=.95\columnwidth]{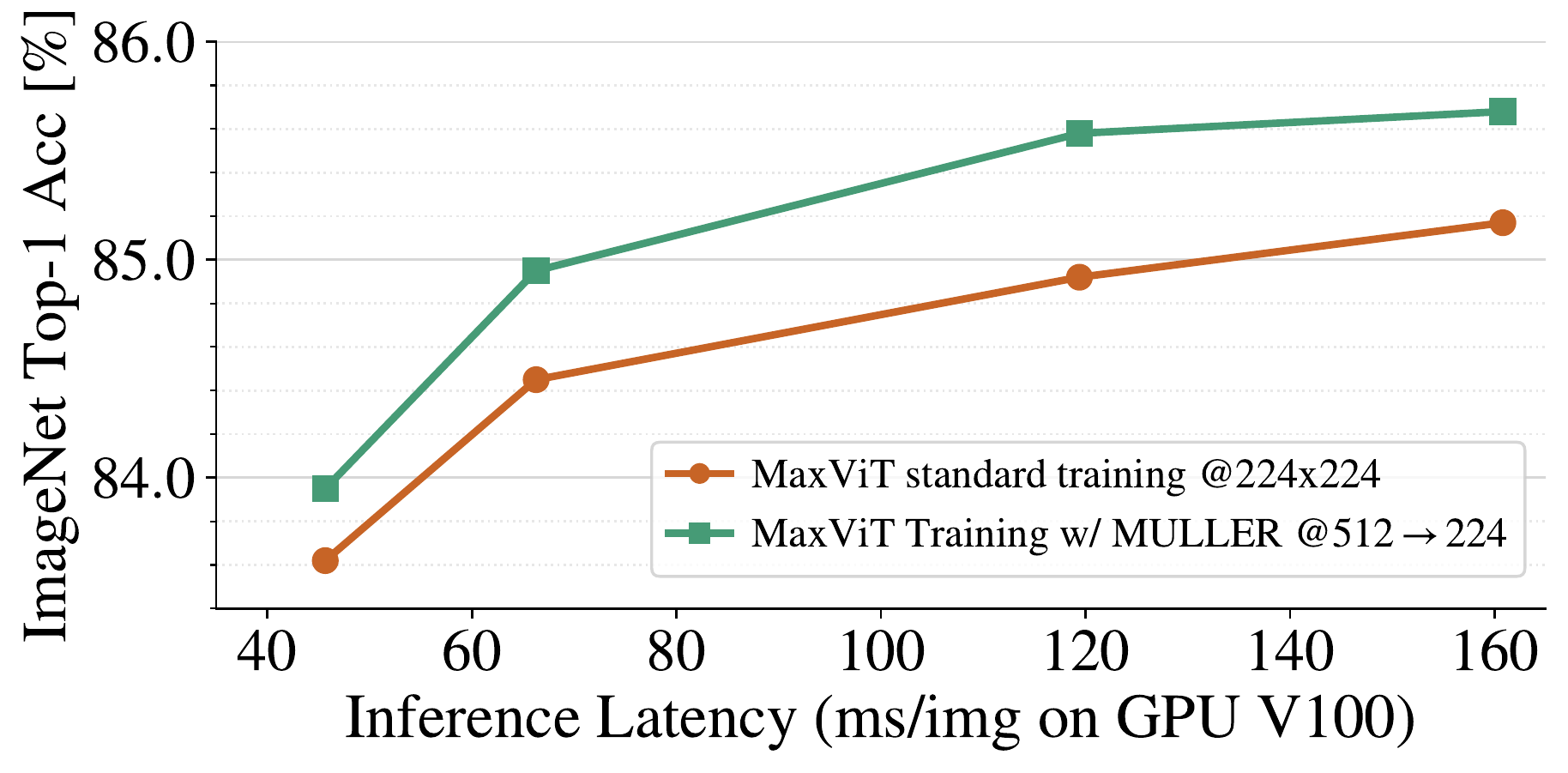} \\
 \includegraphics[width=.95\columnwidth]{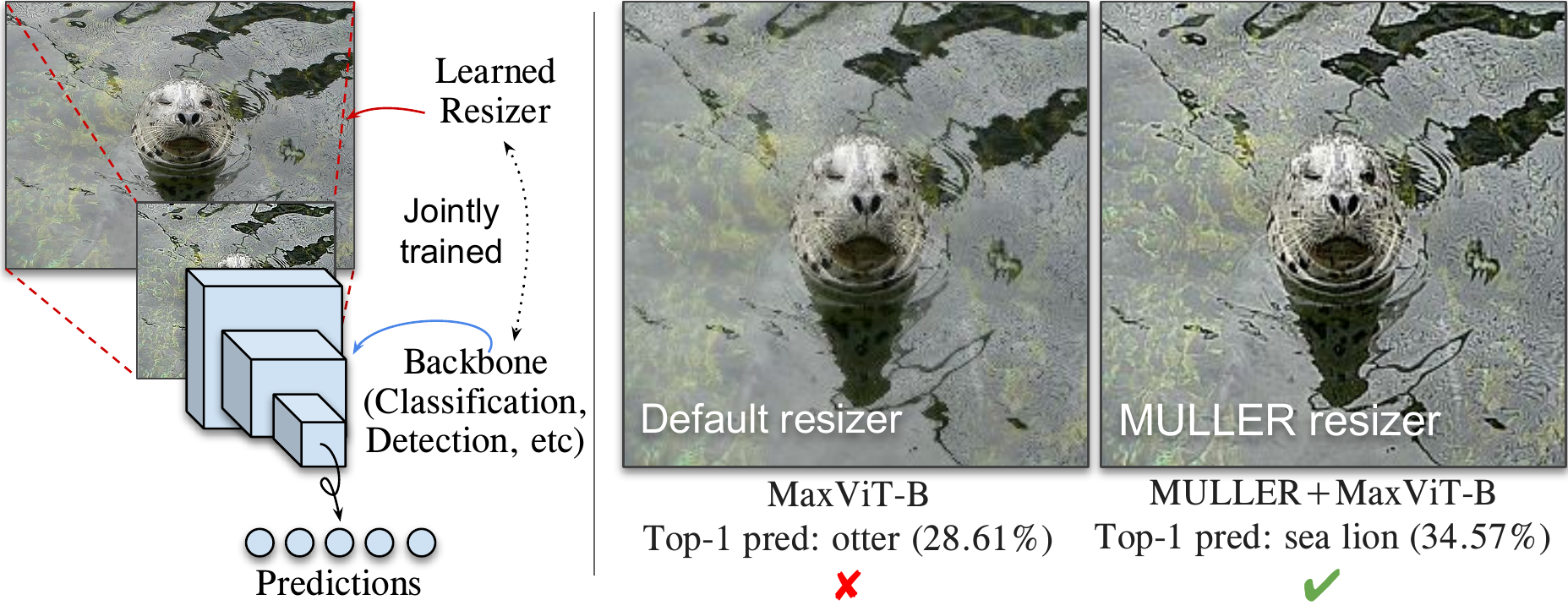}
\end{tabular}
% \vspace{-2mm}
\caption{Top: our proposed learned resizer can push forward a strong vision Transformer MaxViT~\cite{tu2022maxvit} by up to 0.6\% top-1 accuracy on ImageNet-1K with no extra inference cost. Results for other backbones are shown in \cref{ssec:different-backbones}. Bottom: demonstratation of the learned resizer - with detail-boosted input image, the classification accuracy of the examplar image has improved.}
 \label{fig:teaser-fig}
%  \vspace{-3mm}
\end{figure}

Resizing or rescaling refers to the process of changing the resolution of an image, while largely preserving its content for human or machine perception.
There are several major reasons for using resizing:
(1) The mini-batch gradient-based training scheme requires the same image resolution in a batch,
(2) Resizing can help to reduce computational complexity, making it easier and faster to train and inference neural networks,
(3) Smaller images consume lower memory footprint, enabling stable training of large models like Transformers with larger batch-size,
(4) Resizing contributes to improving model generalization and robustness by reducing overfitting to specific image size and scales, making the models more flexible and applicable to real-world scenarios.

Moreover, resizing is an integral component of remote inference frameworks. Typically, to maintain the bandwidth efficiency of the communication network, before sending an image to the inference server, a thumbnail generator down-scales the image to a fixed resolution (\eg 480p). The thumbnail generator can be located on the client side (\eg smart phone), or it can be part of a cloud storage system. This means that in most cases the inference server does not have access to the original image.

Basic resizing functions such as nearest-neighbor or bilinear interpolation have long been the go-to options with little to no deliberate consideration in most training software.
While these simple methods offer greater simplicity and efficiency, they are not optimized for specific computer vision tasks and may lead to the loss of important visual features or details, which can, sometimes, result in significant performance degradation~\cite{parmar2021buggy,talebi2021learning}.
To overcome this limitation, researchers have proposed learned resizers (or downsamplers)~\cite{talebi2021learning, chen2022estimating} that leverage the deep neural networks to learn image resizing directly from data, yielding improved performance on several tasks.
However, one of the main challenges with these learned resizers is that they often require a large number of parameters, and high computational overhead during training and inference.
Note that this is specifically a bottleneck in remote inference where the resizer (a.k.a thumbnail generator) is not in the inference server, and may have limited computational resources to run a heavy neural net resizer. 
Additionally, less-bounded resizers can sometimes be difficult to transfer to new tasks or datasets due to their excessive model capability.

\begin{figure*}[!tb]
\centering
\subfloat{%
  \includegraphics[width=1.85\columnwidth]{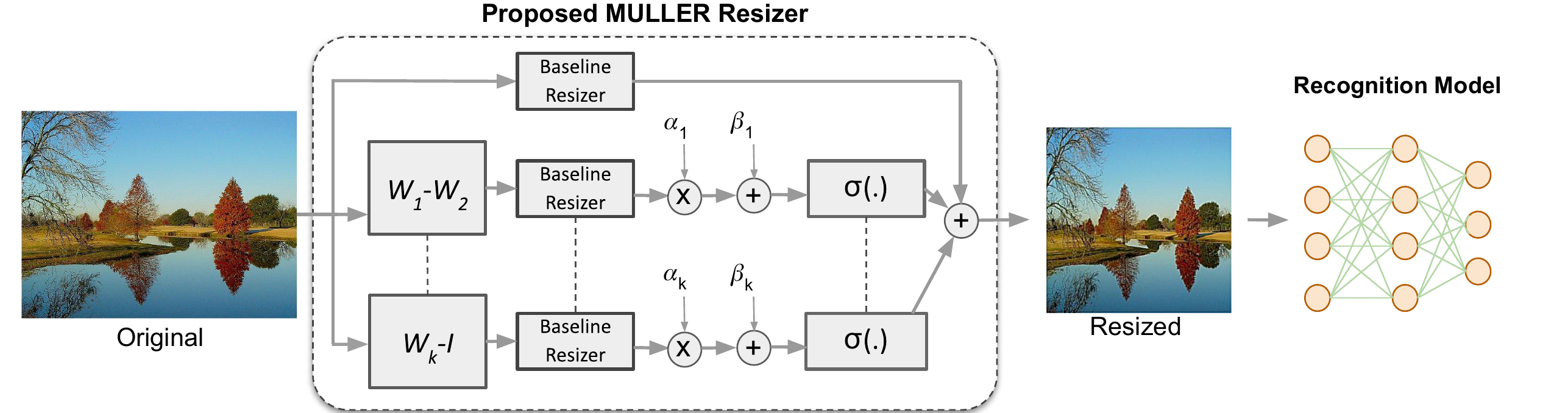}%
}
\caption{\textbf{The architecture of the proposed multilayer Laplacian resizer (MULLER).} The resizer decomposes the input image into multiple layers of Laplacian residuals and then adds them back to the default resized image. The MULLER resizer is jointly trained with the downstream recognition model.}
\label{fig:muller-resizer}
% \vspace{-3mm}
\end{figure*}

In this paper, we introduce an incredibly lightweight learned resizer, which we call MULLER, that operates on multilayer Laplacian decomposition of images (see \cref{fig:muller-resizer}).
Our method requires very few parameters and FLOPs, and does not incur any extra training cost, outperforming existing methods in terms of computational efficiency, parameter efficiency, and transferability.
We show that it is the ability-to-learn that makes a better resizer, but not the capacity of the resizer -- our MULLER resizer only learns four parameters and is more effective than previous complex ones using deep residual blocks~\cite{talebi2021learning}.
We also demonstrate that our method can be used as a drop-in replacement for off-the-shelf resizing functions on several vision tasks, including classification, object detection and segmentation, and image quality assessment, resulting in significant performance improvements without any extra cost.
As shown in \cref{fig:teaser-fig}, for example, training with the MULLER resizer achieves up to 0.6\% performance gain, using a state-of-the-art backbone MaxViT~\cite{tu2022maxvit} as the testbed.
Our contributions are:
\begin{itemize}
% [leftmargin=*]
% \itemsep0em 
    \item We propose a surprisingly simple and lightweight resizer, that can be used as a drop-in replacement for off-the-shelf resizing functions like bilinear resizing.
    % \vspace{-0.6mm}
    \item We demonstrate its applications to multiple computer vision tasks, including image classification, object detection and segmentation, and image quality assessment, showing superior performance over existing approaches.
    \item Extensive ablation studies, analysis, and visualization results are provided to show the robustness and generalization of the proposed resizer for various model scales, benchmarks, and tasks.
    % \vspace{-0.6mm}
\end{itemize}

%------------------------------------------------------------------------
\section{Related Work}

\noindent\textbf{Resizing in vision.}
Resizing is a crucial preprocessing step to train deep learning vision models.
Due to their simplicity, efficiency and availability, nearest-neighbor and bilinear interpolations are the most widely used resizing methods in both training, inference, and serving.
These simple approaches, however, can suffer from detail loss and artifacts, and the degraded image quality might hamper the performance of downstream visual recognition tasks, especially when the resizing factor is large.

Some recent works have explored to use learning-based methods for image downscaling to enhance the desired content in the resized images from training data \cite{chen2021progressive, chen2022estimating, talebi2021learning, zhao2020thumbnet, riad2022learning, li2018learning}.
For example, the authors of \cite{chen2021progressive, chen2022estimating} proposed a residual CNN module for downscaling, and jointly trained it with an image compression network to generate ``compression-friendly'' representations.
\cite{talebi2021learning} introduced a CNN-based learned resizer for various computer vision tasks, including image classification and image quality assessment.
Similarly, the idea of learned rescaling has been applied to other computer vision applications~\cite{li2018learning, riad2022learning, xu2022bridging,zhao2020thumbnet}, showing improved performance in detection and recognition.

\noindent\textbf{Image processing for machine vision.}
Image processing or enhancement problems such as super-resolution~\cite{shi2016real,li2021comisr}, denoising~\cite{zhang2017beyond}, and deblurring~\cite{tao2018scale,tu2022maxim} have been long-standing challenges in computer vision.
Recent works have focused on building larger-scale, diverse benchmarks, exploring novel model architectures, and improving training techniques.
These works aim to produce visually pleasing outputs, often measured by conventional metrics like PSNR or SSIM~\cite{wang2004image}, or through human evaluations, without considering downstream recognition performance of the output.

There exists a number of works relating image processing to image recognition performance, or machine-oriented image processing.
Some works~\cite{zhang2016colorful,sajjadi2016enhancenet,li2019single,li2018benchmarking} use image recognition accuracy as supplementary metrics besides visual quality metrics to evaluate the performance of image restoration models.
Towards recognition-aware training, \cite{haris2021task} proposed to train super-resolution using object detection loss and show promising results over conventional methods.
Other works~\cite{suzuki2019image,talebi2021better} introduce pre-editing networks before image compression to improve compression efficiency without sacrificing classification accuracy.
Recently, Liu et al.~\cite{liu2019transferable} developed an approach to train the processing models under the objective of image recognition accuracy, and investigated the efficacy of popular preprocessing operations such as super-resolution, denoising, and JPEG-deblocking on improving recognition performance.
More similar studies~\cite{sharma2018classification,liu2017image,li2018end,liu2018disentangling} have been conducted to jointly train a processing model like denoising, dehazing, face reconstruction, together with recognition model to achieve better image processing for recognition quality.

\section{Proposed Approach}

In this section, we introduce our proposed multilayer Laplacian resizer (MULLER, overviewed in~\cref{fig:muller-resizer}), and discuss how we employ it for training several popular vision tasks.
Unlike previous proposed resizers~\cite{chen2022estimating,talebi2021learning}, we aim to keep the computational cost of the model as low as possible such that it can replace  existing resizers (e.g., bilinear) without extra cost, but also there is a notable performance gain.
Our proposed approach is different in that (1) it is orders of magnitude faster, hence more scalable (to large image size), (2) it only has a handful of parameters which allows for better generalization, (3) it adds almost no extra training cost to the system.
We show that with learning merely a couple of parameters, training with an added resizing module performs as effective as having a heavy downscaling network with several thousand parameter counts.

\subsection{Resizer Model}

Image resizing models can be generally formulated as:
\begin{equation}
\mathbf{y} = \mathbf{F}_2(\mathbf{R}(\mathbf{F}_1(\mathbf{x}); h', w')),
\end{equation}
where $\mathbf{R}$ maps the input image $\mathbf{x}$ of size $h\times w$ to an output image of size $h'\times w'$ by computing the pixel values at the target spatial locations.
$\mathbf{F}_1$ and $\mathbf{F}_2$ denote optional pre- and post-filtering operations.
Typically, $\mathbf{F}_1$ and $\mathbf{F}_2$ can be identity functions, and $\mathbf{R}$ is chosen as a simple interpolation method like nearest-neighbor, bilinear, or bicubic.
To learn more powerful resizing, learned resizers have been proposed~\cite{chen2021progressive, talebi2021learning, chen2022estimating} by applying a base resizer on intermediate neural activations, wherein $\mathbf{F}_1$ and $\mathbf{F}_2$ are two designed CNNs applied at the original and output resolutions, respectively.
Despite showing promising performance, however, these resizers typically suffer from high computational complexity, and thus their net performance gain might be compromised in terms of the overall inference cost.

\subsection{Proposed MULLER Resizer}

We are inspired by the observations that these different learned resizers, if properly regularized, will often learn to enhance edges, details, or sharpness of the image to benefit downstream tasks~\cite{chen2021progressive, chen2022estimating}.
To this end, we present to date, the simplest learned resizing model, using multilayer Laplacian decomposition, that is able to achieve `bandpassed' detail and texture manipulation with only a handful of learnable parameters.
\cref{fig:muller-resizer} shows the architecture of the proposed MULLER resizer.
MULLER has the following form:
\begin{equation}
\label{eq:muller-eq}
\begin{aligned}
\mathbf{z} = \underbrace{\mathbf{R}(\mathbf{x})}_\text{Base image}+\sum_{\ell=1}^{k}\underbrace{\sigma(\alpha_\ell(\mathbf{R}((\mathbf{W}_\ell-\mathbf{W}_{\ell+1})\mathbf{x})+\beta_\ell)}_\text{Enhanced details in each subband},
\end{aligned}
\end{equation}
where $\mathbf{R}$ denotes the base resizer (\eg bilinear) and $\{\mathbf{W}_1,\mathbf{W}_2,...,\mathbf{W}_k\}$ represents the low-pass filter basis. We define $\mathbf{W}_\ell$ as a positive row-stochastic matrix~\cite{milanfar2012tour} of size $n\times n$, with $n$ representing the number of pixels in the vectorized input image $\mathbf{x}$. Note that we assume $\mathbf{W}_{k+1}=\mathbf{I}$, where $\mathbf{I}$ is the identity matrix. 
Each layer in Eq.~\eqref{eq:muller-eq} (see \cref{fig:muller-resizer}) uses a difference of the filters to decompose the image into different detail layers (bandpass filtering).

Without loss of generality, we choose the Gaussian kernel as our base filter, and generate the filter bank by an iterative application of the same base filter as $\mathbf{W}_\ell=\mathbf{W}^{k-\ell+1}$ with $\mathbf{W}$ being a Gaussian filter with standard deviation 1. Note that the iterative application of the low-pass filter results in a smoother image.
The filtered subband image $(\mathbf{W}_\ell-\mathbf{W}_{\ell+1})\mathbf{x}$ in brach $\ell$ is fed into the base resizer to produce the target resolution layer.
We add trainable scaling and bias parameters $(\alpha_\ell,\beta_\ell)$ per layer to modulate and shift the resized response. Then, a nonlinearity function $\sigma$ (\eg, \texttt{tanh}) is applied on the resulting image layer, and finally the output is added to the base resized image. Note that the scaling factor $\alpha_\ell$ controls the amount of detail boosted or suppressed in layer $l$ of the resizer, and the bias parameter $\beta_\ell$ controls the mean shift.

It is worth pointing out that in this framework, only the scalar and the bias values in the residual layers are trainable, meaning that for $k=3$, there are only six trainable parameters, and the overall computational cost is only applying 4 bilinear resizers and 3 Gaussian filters.
Note that the term ``Laplacian'' refers to an interpretation of the filtering structure in~\cref{fig:muller-resizer} that can be written as a summation of Laplacian operators, namely $\mathbf{L}_\ell=\mathbf{I}-\mathbf{W}_\ell$. More explicitly, for a linear activation, the resulting image $\mathbf{y}$ can be expressed as a Laplacian form~\cite{talebi2016fast}:
\begin{equation}
\mathbf{y}=\gamma_0\mathbf{R}(\mathbf{x})+\gamma_1\mathbf{R}(\mathbf{L}_1\mathbf{x})+...+\gamma_k\mathbf{R}(\mathbf{L}_k\mathbf{x}) + \delta,
\end{equation}

\subsection{Applications in Vision Tasks}

While theoretically the resizer can be a drop-in replacement of the default resizer anywhere in the data generation and machine learning pipeline, we mainly demonstrate its ability to learn more informative thumbnail images for downstream recognition and detection tasks, which account for most of the practical use cases.
We showcase the impact of MULLER by jointly training it with the backbone, where the resizer takes a higher-resolution image from the data pipeline and downscales it to lower-size before feeding as the model inputs.\footnote{Note that MULLER is not limited to downscaling, and in fact should the original image data be low resolution, it can learn to upscale as well.}
Since the proposed resizer is strongly regularized by its design, it needs no extra intermediate loss to train.
The resizer is also task agnostic as no specific changes are needed to train with it in any framework on any vision or even vision-language tasks.
%

%-------------------------------------------------------------------------

\section{Experiments}

We validate the performance of our proposed MULLER resizer on several competitive vision tasks on which resolution plays an important role on performance, including image classification, object detection and segmentation, and image quality assessment.
In order to showcase the impact of MULLER, our main experiments include the state-of-the-art vision Transformer model MaxViT~\cite{tu2022maxvit} as the baseline. We first demonstrate the performance of this baseline model by co-training it with MULLER. Then, we show that MULLER can be effective with other backbones such as ResNet~\cite{he2016deep}, MobileNet-v2~\cite{sandler2018mobilenetv2} and EfficientNet-B0~\cite{tan2019efficientnet}.
In all the experiments, we use 2 layers in MULLER with Gaussian kernel size 5 and standard deviation 1.
We use Tensorflow's default resizer as the base resizer.
More experimental details can be found in \cref{sec:supp-detailed-configs}.

\subsection{Main Experiments on ImageNet Classification}

We demonstrate the efficacy of the MULLER resizer on the standard, but most competitive ImageNet-1K classification task~\cite{krizhevsky2012imagenet}.
We take a top-performing vision Transformer, MaxViT~\cite{tu2022maxvit}, as the backbone model, and pre-train it on ImageNet-1K at 224 $\times$ 224 resolution for 300 epochs.
Instead of directly fine-tuning at higher resolution (\eg, 384 or 512) like previous practices~\cite{liu2021swin, dong2021cswin, dosovitskiy2020image, tu2022maxvit}, we jointly fine-tune the backbone with the MULLER resizer plugged before the stem layers.
We set input and output resolutions as 512 and 224 for MULLER in the ImageNet experiments.

\begin{figure}[!tb]
 \centering
%  \begin{subfigure}[b]{0.48\textwidth}
%      \centering
%      \includegraphics[width=0.99\textwidth]{iccv2023AuthorKit/figures/imagenet_top1_flops.pdf}
%     %  \caption{Accuracy \vs FLOPs performance scaling curve under ImageNet-1K training setting at various input resolution.}
%      \label{fig:imagenet-flops}
%  \end{subfigure}
 \begin{subfigure}[b]{0.47\textwidth}
     \centering
     \includegraphics[width=0.99\textwidth]{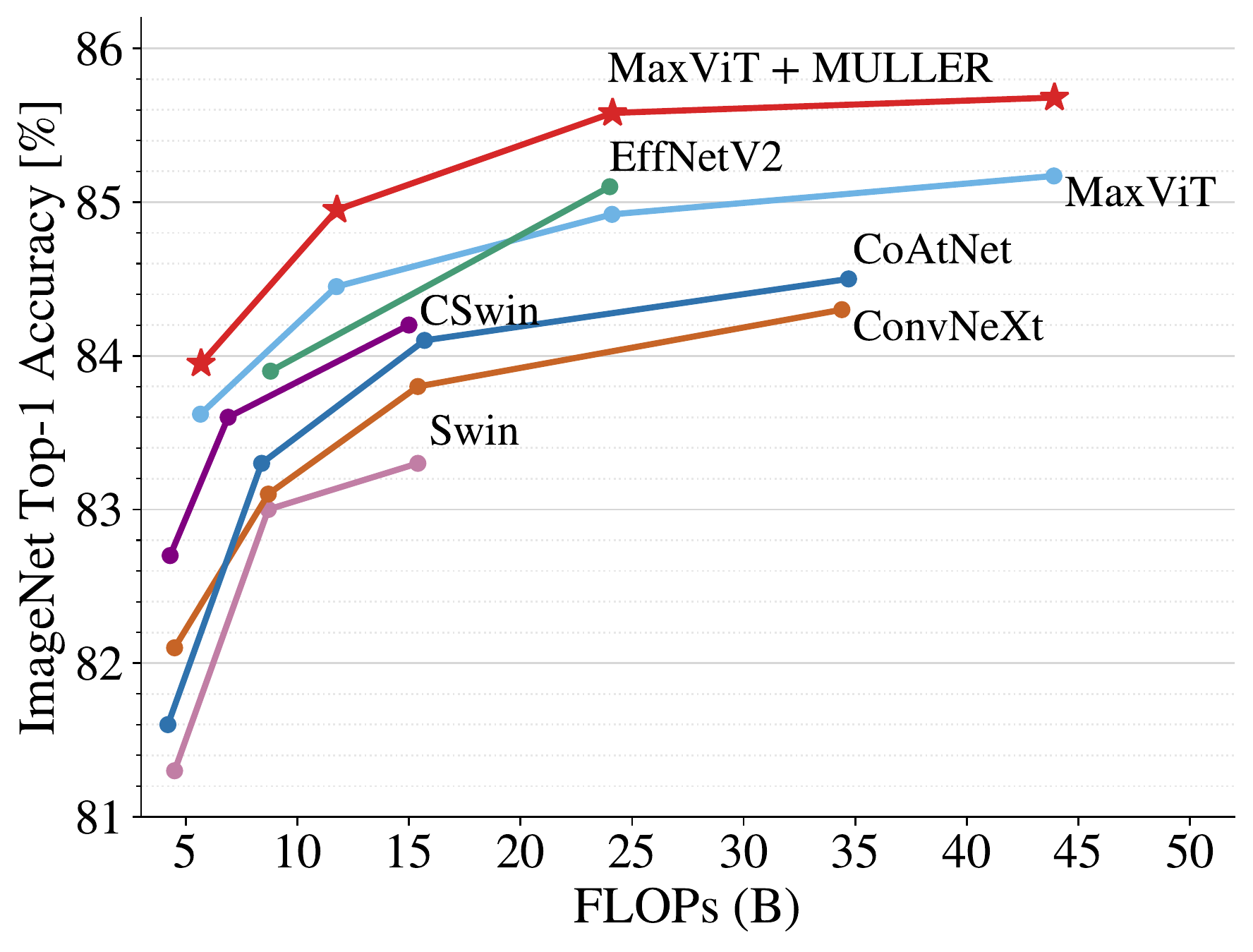}
    %  \caption{Accuracy \vs FLOPs performance scaling curve under ImageNet-1K training setting at various input resolution.}
     \label{fig:imagenet-flops}
 \end{subfigure}

 \hfill
%  \begin{subfigure}[b]{0.48\textwidth}
%      \centering
%      \includegraphics[width=0.99\textwidth]{iccv2023AuthorKit/figures/imagenet_top1_throughput.pdf}
%     %  \caption{Accuracy \vs Parameters scaling curve under ImageNet-1K fine-tuning setting allowing for higher sizes (384/512).}
%      \label{fig:imagenet-thr}
%  \end{subfigure}
 \begin{subfigure}[b]{0.47\textwidth}
     \centering
     \includegraphics[width=0.99\textwidth]{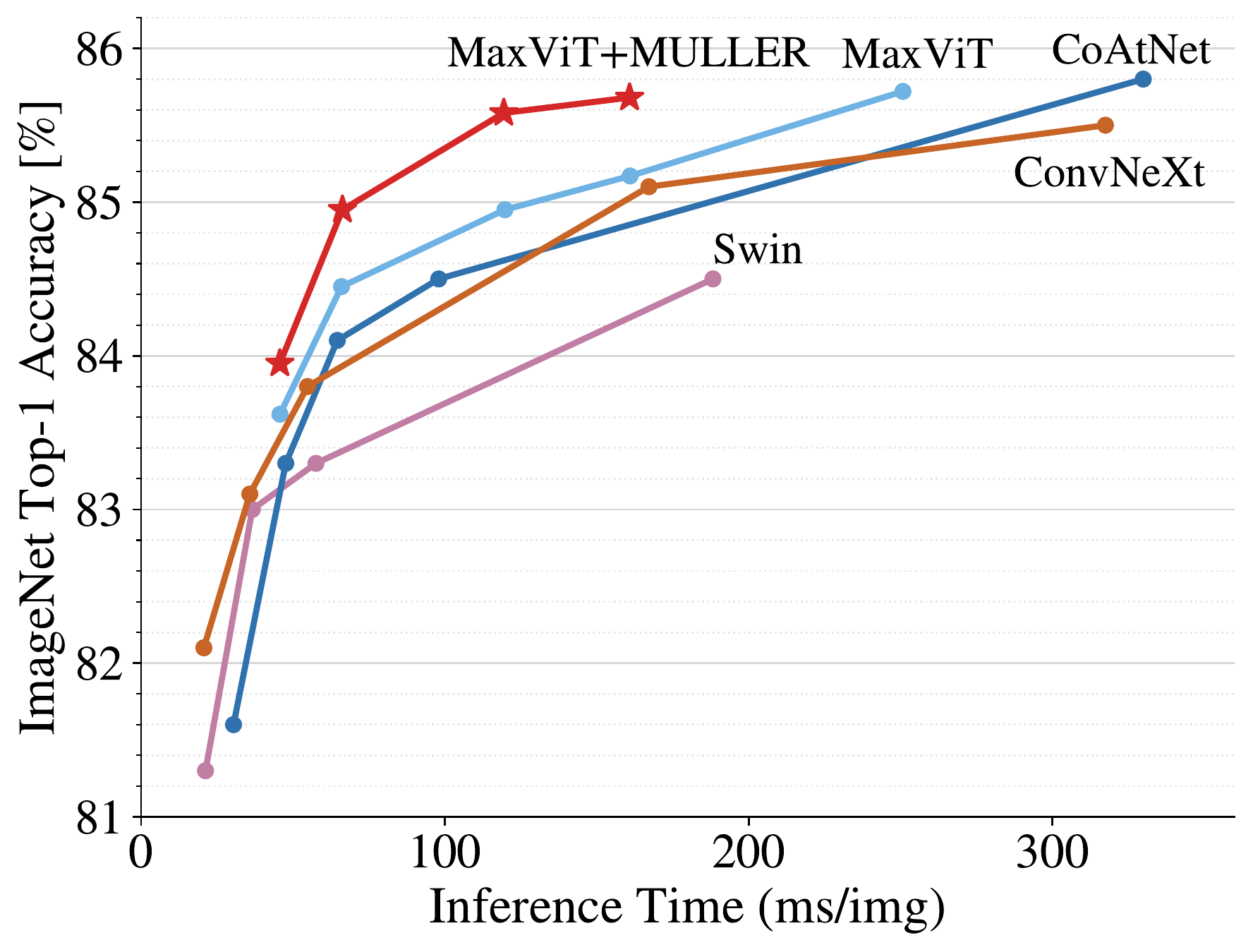}
    %  \caption{Accuracy \vs Parameters scaling curve under ImageNet-1K fine-tuning setting allowing for higher sizes (384/512).}
     \label{fig:imagenet-thr}
 \end{subfigure}
\caption{\textbf{Model FLOPs (top), and Inference Latency (bottom) performance comparison of state-of-the-art vision backbones on ImageNet-1K.} We show that MaxViT trained with MULLER resizer yielded the best accuracy \vs computation and accuracy \vs inference-cost tradeoff. Note that top figure includes only $224\times224$ models whereas the bottom figures include the best possible performance curves among various training size. Inference time is calculated by the throughput in \cref{tab:imagenet1k}.}
\label{fig:imagenet-1k-main}
\vspace{-3mm}
\end{figure}

\noindent\textbf{ImageNet-1K.}
The main results on ImageNet-1K classification are shown in~\cref{tab:imagenet1k} .
Note that we include all the state-of-the-art models trained to their highest possible accuracy reported in the original papers.
For better visualization, we draw the accuracy vs. FLOPs and accuracy vs. inference-latency scaling curves in \cref{fig:imagenet-1k-main}, respectively.
As may be seen, MaxViT powered by the MULLER resizer sets a new state-of-the-art top-1 accuracy 85.68\% with only 43.9B FLOPs among all the compared models trained at 224x224.
MULLER improves at an average of 0.49 accuracy across the four MaxViT variants.
In terms of actual inference time, MaxViT with MULLER exceeds among all the models trained at various resolution -- equivalently, it can save 36\% latency to achieve $\sim85.7\%$ accuracy.

\begin{table}[!t]
\centering
% \footnotesize
\setlength{\tabcolsep}{0.8pt}
\renewcommand{\arraystretch}{1}

\begin{tabular}{l|ccccc}
% \hline
% \rowcolor[gray]{0.95}
\multirow{2}{*}{Model} & \multirow{2}{*}{\begin{tabular}{@{}c@{}}Eval\\size\end{tabular}} & \multirow{2}{*}{Params} & \multirow{2}{*}{FLOPs} & \multirow{2}{*}{\begin{tabular}{@{}c@{}}Thr \\(img/s)\\ \end{tabular}} &
\multirow{2}{*}{\begin{tabular}{@{}c@{}}IN-1K \\top-1 acc\\ \end{tabular}} \\
&&&&\\
% \rowcolor[gray]{0.95}
\toprule
% \multirow{10}{*}{ConvNets} & \textcolor{blueish}{$\bullet$}EffNet-B6~\cite{tan2019efficientnet} & 528  & 43M & 19.0G & 96.9  & 84.0 \\
% \textcolor{blueish}{$\bullet$}EffNet-B7~\cite{tan2019efficientnet} & 600  & 66M & 37.0G  &  55.1  & 84.3 \\
% \textcolor{blueish}{$\bullet$}RegNetY-16~\cite{radosavovic2020designing} & 224 & 84M & 16.0G & 334.7 & 82.9 \\
% \textcolor{blueish}{$\bullet$}NFNet-F0~\cite{brock2021high} & 256 & 72M & 12.4G  & 533.3 & 83.6 \\
% \textcolor{blueish}{$\bullet$}NFNet-F1~\cite{brock2021high} & 320 & 132M & 35.5G  & 228.5 &  84.7 \\
\textcolor{blueish}{$\bullet$}EffNetV2-S~\cite{tan2021efficientnetv2} & 384 & 24M & 8.8B  & 666.6 &  83.9 \\
\textcolor{blueish}{$\bullet$}EffNetV2-M~\cite{tan2021efficientnetv2} & 480 & 55M & 24.0B  & 280.7 &  85.1 \\
\textcolor{blueish}{$\bullet$}ConvNeXt-T~\cite{liu2022convnet} & 224 & 29M & 4.5B & 774 & 82.1\\
\textcolor{blueish}{$\bullet$}ConvNeXt-S~\cite{liu2022convnet} & 224 & 50M & 8.7B  & 447.1 & 83.1 \\
\textcolor{blueish}{$\bullet$}ConvNeXt-B~\cite{liu2022convnet} & 224 & 89M & 15.4B  & 292.1 & 83.8 \\
\textcolor{blueish}{$\bullet$}ConvNeXt-L~\cite{liu2022convnet} & 224 & 198M & 34.4B  & 146.8 & 84.3 \\
\hline
 \textcolor{brickred}{$\circ$}ViT-B/32~\cite{dosovitskiy2020image} & 384 & 86M & 55.4B  &  85.9 & 77.9 \\
\textcolor{brickred}{$\circ$}ViT-B/16~\cite{dosovitskiy2020image} & 384 & 307M & 190.7B & 27.3  & 76.5 \\
% \textcolor{brickred}{$\circ$}DeiT-B~\cite{touvron2021training} & 384 & 86M & 55.4G & 85.9 & 83.1 \\
% \textcolor{brickred}{$\circ$}CaiT-M24~\cite{touvron2021going} & 224 & 186M & 36.0G  & - & 83.4 \\
% \textcolor{brickred}{$\circ$}CaiT-M24~\cite{touvron2021going} & 384 & 186M & 116.1G & - & 84.5 \\
% \textcolor{brickred}{$\circ$}DeepViT-L~\cite{zhou2021deepvit} & 224 & 55M & 12.5G & - & 83.1 \\
% \textcolor{brickred}{$\circ$}T2T-ViT-24~\cite{yuan2021tokens} & 224 & 64M & 15.0G  & - & 82.6 \\
\textcolor{brickred}{$\circ$}Swin-T~\cite{liu2021swin} & 224 & 29M & 4.5B & 755.2 &  81.3 \\
\textcolor{brickred}{$\circ$}Swin-S~\cite{liu2021swin} & 224 & 50M & 8.7B & 436.9 &  83.0 \\
\textcolor{brickred}{$\circ$}Swin-B~\cite{liu2021swin} & 224 & 88M & 15.4B & 278.1 &  83.5 \\
\textcolor{brickred}{$\circ$}CSwin-B~\cite{dong2021cswin} & 224 & 23M & 4.3B & 701 & 82.7 \\
\textcolor{brickred}{$\circ$}CSwin-B~\cite{dong2021cswin} & 224 & 35M & 6.9B & 437 & 83.6 \\
\textcolor{brickred}{$\circ$}CSwin-B~\cite{dong2021cswin} & 224 & 78M & 15.0B & 250 & 84.2 \\
% \textcolor{brickred}{$\circ$}CSwin-B~\cite{dong2021cswin} & 384 & 78M & 47.0B & - & 85.4 \\
% \textcolor{brickred}{$\circ$}Focal-S~\cite{yang2021focal} & 224 & 51M & 9.1G & - & 83.5 \\
% \textcolor{brickred}{$\circ$}Focal-B~\cite{yang2021focal} & 224 & 90M & 16.0G & - & 83.8 \\
\hline
% \textcolor{darkgreen}{$\diamond$}CvT-21~\cite{wu2021cvt} & 384 & 32M & 24.9G  & - & 83.3 \\
\textcolor{darkgreen}{$\diamond$}CoAtNet-0~\cite{dai2021coatnet} & 224 & 25M & 4.2B  & 526 & 81.6 \\
\textcolor{darkgreen}{$\diamond$}CoAtNet-1~\cite{dai2021coatnet} & 224 & 25M & 4.2B  & 336 &  83.3 \\
\textcolor{darkgreen}{$\diamond$}CoAtNet-2~\cite{dai2021coatnet} & 224 & 75M & 15.7B  & 247.7 &  84.1 \\
\textcolor{darkgreen}{$\diamond$}CoAtNet-3~\cite{dai2021coatnet} & 224 & 168M & 34.7B  & 163.3 & 84.5 \\
% \textcolor{darkgreen}{$\diamond$}CoAtNet-3~\cite{dai2021coatnet} & 384 & 168M & 107.4G &  48.5 & 85.8 \\
% \textcolor{darkgreen}{$\diamond$}CoAtNet-3~\cite{dai2021coatnet} & 384 & 168M & 107.4G &  48.5 & 85.8 \\
% \textcolor{darkgreen}{$\diamond$}CoAtNet-3~\cite{dai2021coatnet} & 512 & 168M & 203.1B &  22.4 & 86.0 \\
\textcolor{darkgreen}{$\diamond$}MaxViT-T  & 224 & 31M & 5.6B & 350.4 & 83.62 \\
\rowcolor[gray]{0.95}\textcolor{darkgreen}{$\diamond$}+MULLER$_{512\rightarrow 224}$ & 224 & 31M & 5.63B &  349.6 & 83.95 \\
\textcolor{darkgreen}{$\diamond$}MaxViT-S  & 224 & 69M & 11.7B & 242.5 & 84.45 \\
\rowcolor[gray]{0.95}\textcolor{darkgreen}{$\diamond$}+MULLER$_{512\rightarrow 224}$ & 224 & 69M & 11.73B & 241.4 & 84.95 \\
\textcolor{darkgreen}{$\diamond$}MaxViT-B & 224 & 120M & 23.4B & 133.6 & 84.95 \\
\rowcolor[gray]{0.95}\textcolor{darkgreen}{$\diamond$}+MULLER$_{512\rightarrow 224}$ & 224 & 120M & 23.43B & 133.0 & 85.58 \\
\textcolor{darkgreen}{$\diamond$}MaxViT-L  & 224 &  212M & 43.9B & 99.4 & 85.17 \\
\rowcolor[gray]{0.95}\textcolor{darkgreen}{$\diamond$}+MULLER$_{512\rightarrow 224}$ & 224 &  212M & 43.93B & 99.3 & {85.68} \\
% \textcolor{darkgreen}{$\diamond$}MaxViT-T  & 384 & 31M & 17.7G & 121.9 & 83.62 \\
% \textcolor{darkgreen}{$\diamond$}MaxViT-S & 384 & 69M & 36.1G & 82.7 & 85.24 \\
% \textcolor{darkgreen}{$\diamond$}MaxViT-B & 384 & 120M & 74.2G & 45.8 & 85.74 \\
% \textcolor{darkgreen}{$\diamond$}MaxViT-L & 384 &  212M & 133.1G & 34.3 & 86.34 \\
% \textcolor{darkgreen}{$\diamond$}MaxViT-T & 512 & 31M & 33.7B &  63.8 & 85.72 \\
% % \rowcolor[gray]{0.95}\textcolor{darkgreen}{$\diamond$}+MULLER$_{512\rightarrow 224}$ & 512 & 31M & 5.6B &  350.4 & 83.95 \\
% \textcolor{darkgreen}{$\diamond$}MaxViT-S & 512 & 69M & 67.6B & 43.3 & 86.19 \\
% % \rowcolor[gray]{0.95}\textcolor{darkgreen}{$\diamond$}+MULLER$_{512\rightarrow 224}$ & 512 & 69M & 11.7B & 241.4 & 84.95 \\
% \textcolor{darkgreen}{$\diamond$}MaxViT-B & 512 & 120M & 138.5B & 24.0 & 86.66 \\
% % \rowcolor[gray]{0.95}\textcolor{darkgreen}{$\diamond$}+MULLER$_{512\rightarrow 224}$ & 512 & 120M & 23.4B & 134.0 & 86.37 \\
% \textcolor{darkgreen}{$\diamond$}MaxViT-L & 512 &  212M & 245.4B & 17.8 & {86.70} \\
% \rowcolor[gray]{0.95}\textcolor{darkgreen}{$\diamond$}+MULLER$_{512\rightarrow 224}$ & 512 &  212M & 43.9B & 99.5 & {86.73} \\
% \hline
% \bottomrule
\end{tabular}
\caption{\textbf{Performance comparison under the ImageNet-1K setting.} MULLER$_{\text{A}\rightarrow\text{B}}$ denote that MULLER resizes from A to B, where the backbone takes images of size B. FLOPs counts the total computation of the resizer and backbone. Throughput (Thr) is measured on a single V100 GPU with batch size 16, following~\cite{liu2021swin,tan2021efficientnetv2,liu2022convnet}. \textcolor{blueish}{$\bullet$}, \textcolor{brickred}{$\circ$}, and \textcolor{darkgreen}{$\diamond$} denote ConvNets, Transformers, and hybrid models, respectively.}
\label{tab:imagenet1k}
\vspace{-3mm}
\end{table}

\noindent\textbf{ImageNet-21K and JFT.}
To demonstrate the scaling properties of the MULLER resizer with respect to data size, in~\cref{tab:imagenet21k-jft-comparison} we report the results of the models pre-trained on ImageNet-21K and JFT-300M~\cite{sun2017revisiting}, respectively. 
It can be seen that with ImageNet-21k pretraining, the finetuned MaxViT with MULLER$_{512\rightarrow 224}$ gains 0.8\%, 0.6\%, and 0.7\% accuracy over directly finetuning without resizer for B, L, and XL models, respectively.
Similarly for JFT-300M pretraining, those numbers are 0.8\%, 0.7\%, and 0.7\%.
It indicates that when finetuning with MULLER, MaxViT scales consistently when data size increases from ImageNet-1k up to JFT-300M.

We further observe that for larger models and larger training sets, the backbone can benefit even more through seeing larger input images.
Thus, we also report the performance of training with MULLER$_{576\rightarrow 288}$.
We can see that it further boosts the performance by an average of 0.4$\sim$0.5\% across the board for both 21K and JFT.
Remarkably, MaxViT-XL with MULLER$_{576\rightarrow 288}$ achieves 89.16\% top-1 accuracy with only 162.9B FLOPs.

We also examine the generalization of the resizer across different model variants.
We found that the learned weights in MULLER are very close across different variants, and the transferring results are as effective as the original training.
Detailed results are in the \cref{sec:supp-additional-results}. Note that we present our generalization experiments across different backbones in the next section.

\begin{table}[!tb]
% \footnotesize
\centering
\setlength{\tabcolsep}{0.8pt}
\renewcommand{\arraystretch}{1}

\begin{tabular}{@{}l|ccccc@{}}
% \hline
% \rowcolor[gray]{0.95}
\multirow{2}{*}{Model} & \multirow{2}{*}{\begin{tabular}{@{}c@{}}Eval\\size\end{tabular}} & \multirow{2}{*}{Params} & \multirow{2}{*}{FLOPs} &
\multicolumn{2}{c}{IN-1K top-1 acc.} \\\cline{5-6}
% \vspace{0.5mm}
\rule{0pt}{2ex}& & & &  21K-pt & JFT-pt\\ 
% \rowcolor[gray]{0.95}
\toprule

\textcolor{blueish}{$\bullet$}BiT-R-101x3~\cite{kolesnikov2020big} & 384 & 388M & 204.6B & 84.4 & - \\
\textcolor{blueish}{$\bullet$}BiT-R-152x4~\cite{kolesnikov2020big} & 480 & 937M & 840.5B & 85.4 & -\\
\textcolor{blueish}{$\bullet$}EffNetV2-L~\cite{tan2021efficientnetv2} & 480 & 121M & 53.0B  &  86.8 & - \\
\textcolor{blueish}{$\bullet$}EffNetV2-XL~\cite{tan2021efficientnetv2} & 512 & 208M & 94.0B &  87.3 & - \\
\textcolor{blueish}{$\bullet$}ConvNeXt-L~\cite{liu2022convnet} & 384 & 198M & 101.0B &  87.5 & - \\
\textcolor{blueish}{$\bullet$}ConvNeXt-XL~\cite{liu2022convnet} & 384 & 350M & 179.0B & 87.8 & -  \\
\textcolor{blueish}{$\bullet$}NFNet-F4+~\cite{brock2021high} & 512 & 527M & 367B & - & 89.20  \\
\hline

\textcolor{brickred}{$\circ$}ViT-B/16~\cite{dosovitskiy2020image} & 384 & 87M & 55.5B & 84.0 & - \\
\textcolor{brickred}{$\circ$}ViT-L/16~\cite{dosovitskiy2020image} & 384 & 305M & 191.1B &  85.2 \\
\textcolor{brickred}{$\circ$}ViT-L/16~\cite{dosovitskiy2020image} & 512 & 305M & 364B &  - & 87.76 \\
\textcolor{brickred}{$\circ$}ViT-H/14~\cite{dosovitskiy2020image} & 518 & 632M & 1021B &  - & 88.55 \\
\textcolor{brickred}{$\circ$}HaloNet-H4~\cite{vaswani2021scaling} & 512 & 85M & - &  85.8 & - \\
\textcolor{brickred}{$\circ$}SwinV2-B~\cite{liu2021swin} & 384 & 88M & - &  87.1 & - \\
\textcolor{brickred}{$\circ$}SwinV2-L~\cite{liu2021swin} & 384 & 197M & -& 87.7 & - \\
\hline

\textcolor{darkgreen}{$\diamond$}CvT-W24~\cite{wu2021cvt} & 384 & 277M & 193.2B &  87.7 & - \\
\textcolor{darkgreen}{$\diamond$}R+ViT-L/16~\cite{dosovitskiy2020image} & 384 & 330M & - & - & 87.12 \\
\textcolor{darkgreen}{$\diamond$}CoAtNet-3~\cite{dai2021coatnet} & 384 & 168M & 107.4B &  87.6 & 88.52 \\
\textcolor{darkgreen}{$\diamond$}CoAtNet-3~\cite{dai2021coatnet} & 512 & 168M & 214B &  87.9 & 88.81 \\
\textcolor{darkgreen}{$\diamond$}CoAtNet-4~\cite{dai2021coatnet} & 512 & 275M & 360.9B &  88.1 & 89.11 \\
% \textcolor{darkgreen}{$\diamond$}MaxViT-B & 384 & 119M & 74.2B  & 88.24 & 88.69 \\
\textcolor{darkgreen}{$\diamond$}MaxViT-B & 224 & 119M & 23.4B  & 86.63 & 87.05 \\
\rowcolor[gray]{0.95}\textcolor{darkgreen}{$\diamond$}+MULLER$_{512\rightarrow 224}$ & 224  & 119M & 23.4B & 87.40 & 87.82 \\
\rowcolor[gray]{0.95}\textcolor{darkgreen}{$\diamond$}+MULLER$_{576\rightarrow 288}$ & 288 & 119M & 40.6B & 87.92 & 88.39 \\
\textcolor{darkgreen}{$\diamond$}MaxViT-L & 224 & 212M & 43.9B & 86.86 & 87.72 \\
\rowcolor[gray]{0.95}\textcolor{darkgreen}{$\diamond$}+MULLER$_{512\rightarrow 224}$ & 224 & 212M & 43.9B & 87.48 &  88.43  \\
\rowcolor[gray]{0.95}\textcolor{darkgreen}{$\diamond$}+MULLER$_{576\rightarrow 288}$ & 288 & 212M & 73.4B & 87.94 &  88.87  \\
\textcolor{darkgreen}{$\diamond$}MaxViT-XL & 224 & 475M & 97.8B & 87.25 &  88.06 \\
\rowcolor[gray]{0.95}\textcolor{darkgreen}{$\diamond$}+MULLER$_{512\rightarrow 224}$ & 224 & 475M & 97.8B & 87.90 & 88.74 \\
\rowcolor[gray]{0.95}\textcolor{darkgreen}{$\diamond$}+MULLER$_{576\rightarrow 288}$ & 288 & 475M & 162.9B & 88.31 & 89.16 \\
\textcolor{darkgreen}{$\diamond$}MaxViT-B & 512 & 119M & 138.3B & 88.38 &  88.82 \\
\textcolor{darkgreen}{$\diamond$}MaxViT-L & 512 &  212M & 245.2B & 88.46 &  89.41 \\
\textcolor{darkgreen}{$\diamond$}MaxViT-XL & 512 & 475M & 535.2B & {88.70} & {89.53}  \\
% \bottomrule
\end{tabular}
\caption{\textbf{Performance comparison for large-scale data regimes}: ImageNet-21K and JFT pretrained models. We report results using two different settings: MULLER$_{512\rightarrow 224}$ and MULLER$_{576\rightarrow 288}$ respectively, as we observe that on larger models and larger training sets, the backbone benefits more by seeing larger inputs.}
\label{tab:imagenet21k-jft-comparison}
\vspace{-3mm}
\end{table}%

\subsection{Different Backbones}
\label{ssec:different-backbones}

\noindent\textbf{Main results.}
To explore the resizer beyond the MaxViT architecture, we selected some widely used backbones including ResNet-50~\cite{he2016deep}, EfficientNet-B0~\cite{tan2019efficientnet}, and MobileNet-v2~\cite{sandler2018mobilenetv2}.
Our results are presented in \cref{tab:imagenet1k-backbones}. We also make comparisons with the resizer of Talebi et al.~\cite{talebi2021learning}.
We observed that the proposed resizer improves the performance of the baseline backbones consistently.
Also compared to \cite{tan2019efficientnet}, MULLER requires a significantly lower number of FLOPs (two orders-of-magnitude), and in some cases such as MobileNet-v2 and EfficientNet-B0 it outperforms \cite{talebi2021learning}.
These results also indicate that MULLER improves over baseline resizers low the FLOPs regime as well. 

\noindent\textbf{Cross-model Generalization.}
In order to examine the generalizability of MULLER, we evaluate classification models with resizers that are trained with other backbones.
To this end, we first present the learned resizer parameters for each backbone, and then discuss the classification performances.

Results in \cref{tab:imagenet1k-backbones-params} represent the learned MULLER parameters (see Eq. \ref{eq:muller-eq}) for each backbone model trained on ImageNet-1k.
We observed that (1) performance of the classification models are more sensitive to $\alpha_1$ than $\alpha_2$, and (2) the learned bias values are relatively small, meaning the resizer does not significantly shift the mean of each residual image layer.
Note that $|\alpha_\ell| > 1$ means the image details represented by the $\ell$-th layer are boosted, whereas $|\alpha_\ell| < 1$ has the opposite effect. 

To quantify generalizability of the resizer, we used the learned parameters in \cref{tab:imagenet1k-backbones-params} to evaluate different backbones.
As for different backbones, \cref{tab:imagenet1k-cross-validation-backbone} shows that one model leads to classification performance that is in the average proximity of $0.15$ from the best top-1 accuracy.
We believe this can be explained by the fact that MULLER is a constrained model with only 4 trainable parameters. Also, it is important to highlight that in contrast to the resizer in \cite{talebi2021learning}, MULLER does not require fine-tuning.

\noindent\textbf{Impact of Aliasing.} 
It has been shown that aliasing may impact the performance of some deep vision models~\cite{vasconcelos2021impact,parmar2021buggy}.
It is worth mentioning that the results presented in this section are based on anti-aliased images.
More specifically, we used the \texttt{AREA} downscaling method in TensorFlow to produce $512^2$ inputs to MULLER.
We observed that while removing anti-aliasing does not hamper the overall performance gain obtained by MULLER, the learned parameters may differ from~\cref{tab:imagenet1k-backbones-params}.
We will present our results without anti-aliasing in \cref{sec:supp-anti-alias}. 

\begin{table}[!t]
\centering
% \footnotesize
\setlength{\tabcolsep}{4pt}
\renewcommand{\arraystretch}{1}

\begin{tabular}{l|cccc}
% \hline
% \rowcolor[gray]{0.95}
\multirow{1}{*}{Model} & \multirow{1}{*}{\begin{tabular}{@{}c@{}}Size\end{tabular}} & \multirow{1}{*}{FLOPs} &
\multirow{1}{*}{\begin{tabular}{@{}c@{}}top-1 acc.\\ \end{tabular}} \\
% &&&&\\
% \rowcolor[gray]{0.95}
\toprule

EffNet-B0~\cite{tan2019efficientnet} & 224 & 0.39B & 77.1  \\
+\cite{talebi2021learning}$_{512\rightarrow 224}$ & 224 & 2.63B & 77.9  \\
\rowcolor[gray]{0.95}+MULLER$_{512\rightarrow 224}$ & 224 & 0.42B & 78.2  \\

MobileNet-v2~\cite{sandler2018mobilenetv2} & 224 & 0.60B &  70.5 \\
+\cite{talebi2021learning}$_{512\rightarrow 224}$ & 224 & 2.84B & 71.5  \\
\rowcolor[gray]{0.95}+MULLER$_{512\rightarrow 224}$ & 224 & 0.63B & 71.8  \\

ResNet-50~\cite{he2016deep} & 224 & 6.97B & 75.3 \\
+\cite{talebi2021learning}$_{512\rightarrow 224}$ & 224 & 9.33B & 76.2  \\
\rowcolor[gray]{0.95}+MULLER$_{512\rightarrow 224}$ & 224 & 7.0B & 76.2  \\

% \hline
% \bottomrule
\end{tabular}
\caption{\normalsize \textbf{Performance comparison under ImageNet-1K setting with different backbones.}}
\label{tab:imagenet1k-backbones}
\end{table}

\begin{table}[!t]
\centering
% \footnotesize
\setlength{\tabcolsep}{4pt}
\renewcommand{\arraystretch}{1}

\begin{tabular}{l|ccccc}
% \hline
% \rowcolor[gray]{0.95}
\multirow{1}{*}{Model} & \multirow{1}{*}{\begin{tabular}{@{}c@{}}$\alpha_1$\end{tabular}} & \multirow{1}{*}{$\beta_1$} &
\multirow{1}{*}{\begin{tabular}{@{}c@{}}$\alpha_2$\\ \end{tabular}} &
\multirow{1}{*}{$\beta_2$} \\
% &&&&\\
% \rowcolor[gray]{0.95}
\toprule
% MaxViT-T~\cite{tu2022maxvit} & 0.310 & 0.021 & -0.738 & 0.020 \\
% MaxViT-S~\cite{tu2022maxvit} & 0.359 & -0.003 & -0.779 & -0.003 \\
% MaxViT-B~\cite{tu2022maxvit} & 1.022 & -0.037 & -1.185 & -0.026  \\
% MaxViT-L~\cite{tu2022maxvit} & 1.000 & -0.015 & -1.122 & -0.006 \\
% \hline

EffNet-B0~\cite{tan2019efficientnet} & 1.715 & 0.088 & -8.41 & 0.001 \\
MobileNet-v2~\cite{sandler2018mobilenetv2} & 1.480 & 0.174 & -5.25 & -0.058 \\
ResNet-50~\cite{he2016deep} & 1.892 & -0.014 & -11.295 & 0.003 \\

% \hline
% \bottomrule
\end{tabular}
\caption{\normalsize \textbf{The learned MULLER parameters for different backbone models train on ImageNet-1k.}}
\label{tab:imagenet1k-backbones-params}
\end{table}
\begin{table}[!t]
\centering
% \footnotesize
\setlength{\tabcolsep}{2pt}
\renewcommand{\arraystretch}{1}

\begin{tabular}{l|ccc}
% \hline
% \rowcolor[gray]{0.95}
\multirow{1}{*}{Model} & \multirow{1}{*}{\begin{tabular}{@{}c@{}}EffNet-B0\end{tabular}} & \multirow{1}{*}{MobileNet-v2} &
\multirow{1}{*}{\begin{tabular}{@{}c@{}}ResNet-50\\ \end{tabular}} \\
% \rowcolor[gray]{0.95}
\toprule

MULLER$_\text{EffNet}$ & 78.2 & 71.6 &  75.9 \\
MULLER$_\text{MobileNet}$ & 78.0 & 71.8 & 76.0 \\
MULLER$_\text{ResNet}$ & 78.1 & 71.7 & 76.2  \\
% \hline
% \bottomrule
\end{tabular}
\caption{\normalsize \textbf{Cross-model validation of the MULLER resizer for ImageNet-1K on different backbones.}}
\label{tab:imagenet1k-cross-validation-backbone}
\vspace{-3mm}
\end{table}

% \subsection{Different Backbones}
\subsection{Downstream tasks}

\noindent\textbf{Object Detection and Instance Segmentation.}
We evaluated the performance of MULLER on COCO2017~\cite{lin2014microsoft} for object bounding box detection and instance segmentation tasks with a two-stage cascaded Mask-RCNN framework~\cite{ren2015fasterrcnn}.
We warm start the MaxViT backbone using checkpoints pretrained on ImageNet-1K, then finetune the whole model including the resizer on COCO.

\cref{tab:coco} summarizes the object detection and instance segmentation results comparing state-of-the-art ConvNets and vision Transformers.
AP and AP$^m$ denote box and mask average precision.
We report the train and evaluation resolutions as well as their corresponding FLOPs as reference for model complexity.
It may be seen that MaxViT suffers from noticeable performance drop if training resolution is lower.
However, we observed that training with the MULLER resizer can further improve the performance across the board.
Specifically, on MaxViT-B at $640\times 640$, finetuning with MULLER gains 0.7 AP and 0.6 mask AP on the COCO validation set without any FLOPs overhead.

\begin{table}[!t]
\centering
\footnotesize
\setlength{\tabcolsep}{0.8pt}
\renewcommand{\arraystretch}{1.2}
\begin{tabular}{@{}l|ccccccc|cc@{}}
Backbone & Resolution & AP & AP$_{50}$ & AP$_{75}$ & AP$^{m}$ & AP$^{m}_{50}$ & AP$^{m}_{75}$ & FLOPs  \\
\toprule
\textcolor{blueish}{$\bullet$}ResNet-50~\cite{he2016deep} & 1280$\times$800 & 46.3 & 64.3 & 50.5 & 40.1 & 61.7 & 43.4 & 739B  \\
\textcolor{blueish}{$\bullet$}X101-32~\cite{xie2017aggregated} & 1280$\times$800 & 48.1 & 66.5 & 52.4 & 41.6 & 63.9 & 45.2 & 819B \\
\textcolor{blueish}{$\bullet$}X101-64~\cite{xie2017aggregated} & 1280$\times$800 & 48.3 & 66.4 & 52.3 & 41.7 & 64.0 & 45.1 & 972B \\
\textcolor{blueish}{$\bullet$}ConvNeXt-T~\cite{liu2022convnet} & 1280$\times$800 & 50.4 & 69.1 & 54.8 & 43.7 & 66.5 & 47.3 &741B   \\
\textcolor{blueish}{$\bullet$}ConvNeXt-S~\cite{liu2022convnet} & 1280$\times$800 & 51.9 & 70.8 & 56.5 & 45.0 & 68.4 & 49.1 & 827B  \\
\textcolor{blueish}{$\bullet$}ConvNeXt-B~\cite{liu2022convnet} & 1280$\times$800 & 52.7 & 71.3 & 57.2 & 45.6 & 68.9 & 49.5 & 964B \\
%\textcolor{blueish}{$\bullet$}ConvNeXt-L$^\ddagger$ & 1280$\times$800 & 54.8 & 73.8 & 59.8 & 47.6 & 71.3 & 51.7 & 1354 & -\\
\hline
\textcolor{brickred}{$\circ$}Swin-T~\cite{liu2021swin} & 1280$\times$800 &  50.4 & 69.2 & 54.7 & 43.7 & 66.6 & 47.3 & 745B  \\
\textcolor{brickred}{$\circ$}Swin-S~\cite{liu2021swin} & 1280$\times$800 & 51.9 & 70.7 & 56.3 & 45.0 & 68.2 & 48.8 & 838B  \\
\textcolor{brickred}{$\circ$}Swin-B~\cite{liu2021swin} & 1280$\times$800 & 51.9 & 70.5 & 56.4 & 45.0 & 68.1 & 48.9 & 982B  \\
%\textcolor{brickred}{$\circ$}Swin-L$^\ddagger$ & 1280$\times$800 & 53.9 & 72.4 & 58.8 & 46.7 & 70.1 & 50.8 & 1382 & - \\
%\textcolor{brickred}{$\circ$}Swin-L$^\ddagger$ & 1280$\times$800 & 53.9 & 72.4 & 58.8 & 46.7 & 70.1 & 50.8 & 1382 & - \\
\textcolor{brickred}{$\circ$}UViT-T~\cite{chen2022uvit} & 896$\times$896 & 51.1 & 70.4 & 56.2 & 43.6 & 67.7 & 47.2 & 613B  \\ % https://tensorboard.corp.google.com/experiment/2533535615840984763/#timeseries
\textcolor{brickred}{$\circ$}UViT-S~\cite{chen2022uvit} & 896$\times$896 & 51.4 & 70.8 & 56.2 & 44.1 & 68.2 & 48.0 & 744B \\ %https://tensorboard.corp.google.com/experiment/2187330177719871319/#timeseries
\textcolor{brickred}{$\circ$}UViT-B~\cite{chen2022uvit} & 896$\times$896 & 52.5 & 72.0 & 57.6 & 44.3 & 68.7 & 48.3 & 975B  \\ % https://tensorboard.corp.google.com/experiment/4467822036480772036/#timeseries
% \textcolor{brickred}{$\circ$}As-ViT-L~\cite{chen2022autoscaling} & 1024$\times$1024 & 52.7 & 72.3 & 57.9 & 45.2 & 69.7 & 49.8 & 1094G  \\ %https://tensorboard.corp.google.com/experiment/7624284926121512164/#timeseries

\hline
\textcolor{darkgreen}{$\diamond$}MaxViT-T & 640$\times$640  & 49.9 & 69.9 & 54.6 & 42.7 & 66.6 & 46.4 & 379B \\
\rowcolor[gray]{0.95}\textcolor{darkgreen}{$\diamond$}+MULR$_{896\rightarrow640}$ & 640$\times$640 & 50.5 & 70.7 & 55.0 & 43.1 & 67.0 & 46.7 & 379B \\
\textcolor{darkgreen}{$\diamond$}MaxViT-S & 640$\times$640 & 50.5 & 70.2 & 55.3 & 43.3 & 67.3 & 46.8 & 432B \\
\rowcolor[gray]{0.95}\textcolor{darkgreen}{$\diamond$}+MULR$_{896\rightarrow640}$ & 640$\times$640 & 50.8 & 70.4 & 55.5 & 43.5 & 67.7 & 47.1 & 432B  \\
\textcolor{darkgreen}{$\diamond$}MaxViT-B & 640$\times$640 & 51.6 & 71.3 & 56.1 & 44.1 & 68.5 & 47.7 & 543B \\
\rowcolor[gray]{0.95}\textcolor{darkgreen}{$\diamond$}+MULR$_{896\rightarrow640}$ & 640$\times$640 & 52.3 & 71.5 & 57.0 & 44.7 & 68.9 & 48.7 & 543B \\
\textcolor{darkgreen}{$\diamond$}MaxViT-T & 896$\times$896 & 52.1 & 71.9 & 56.8 & 44.6 & 69.1 & 48.4 & 475B \\
\textcolor{darkgreen}{$\diamond$}MaxViT-S & 896$\times$896 & 53.1 & 72.5 & 58.1 & 45.4 & 69.8 & 49.5 & 595B \\
\textcolor{darkgreen}{$\diamond$}MaxViT-B & 896$\times$896 & {53.4} & {72.9} & {58.1} & {45.7} & {70.3} & {50.0} & 856B \\
% \bottomrule
\end{tabular}
\caption{\textbf{Comparison of two-stage object detection and instance segmentation on COCO2017.}
All models are pretrained on ImageNet-1K.}
\label{tab:coco}
% \vspace{-1mm}
\end{table}
\begin{table}[!t]
\centering
% \centeringbat
\setlength{\tabcolsep}{1.6pt}
\renewcommand{\arraystretch}{1.0}

\begin{tabular}{l|cccc}
% \hline
% \rowcolor[gray]{0.95}
Model & Size & Params & PLCC$\uparrow$ \\
\toprule
\textcolor{blueish}{$\bullet$}NIMA~\cite{talebi2018nima} & $224$ & 56M & 0.636  \\
\textcolor{blueish}{$\bullet$}+\cite{talebi2021learning}$_{512\rightarrow 224}$ & $224$ & 56M & 0.680 \\
\textcolor{blueish}{$\bullet$}EffNet-B0~\cite{tan2019efficientnet} & $224$ & 5.3M & 0.642 \\
\textcolor{blueish}{$\bullet$}+\cite{talebi2021learning}$_{512\rightarrow 224}$ & $224$ & 5.3M & 0.650 \\
\textcolor{blueish}{$\bullet$}AFDC\cite{chen2020adaptive} & $224$ & 44.5M & 0.671 \\\hline
\textcolor{brickred}{$\circ$}ViT-S/32~\cite{ke2021musiq} & $384$ & 22M & 0.665 \\
\textcolor{brickred}{$\circ$}ViT-B/32~\cite{ke2021musiq} & $384$ & 88M & 0.664  \\
\textcolor{brickred}{$\circ$}MUSIQ~\cite{ke2021musiq} & \footnotesize{$224\!\sim\!512$} & 27M & 0.720  \\\hline
%\textcolor{brickred}{$\circ$}MUSIQ~\cite{ke2021musiq} & Full-res & 27M & 0.731 & \textbf{0.719} \\\hline
\textcolor{darkgreen}{$\diamond$}MaxViT-T & $224$ & 31M & 0.707 \\
\rowcolor[gray]{0.95}\textcolor{darkgreen}{$\diamond$}+MULLER$_{512\rightarrow 224}$ & $224$ & 31M & \textbf{0.729} \\
%\textcolor{darkgreen}{$\diamond$}MaxViT-T & $384$ & 31M & 0.736 & 0.699 \\
%\textcolor{darkgreen}{$\diamond$}MaxViT-T & $512$ & 31M & \textbf{0.745} & \textbf{0.708} \\
% \bottomrule
\end{tabular}
\caption{\textbf{Image aesthetic assessment results on the AVA benchmark~\cite{murray2012ava}}. PLCC represents the Pearson's linear correlation coefficient.}
\label{tab:iqa-comparison}
% \vspace{-3mm}
\end{table}

\begin{figure*}[!tb]
 \centering
 \begin{subfigure}[b]{0.96\textwidth}
     \centering
     \includegraphics[width=0.96\textwidth]{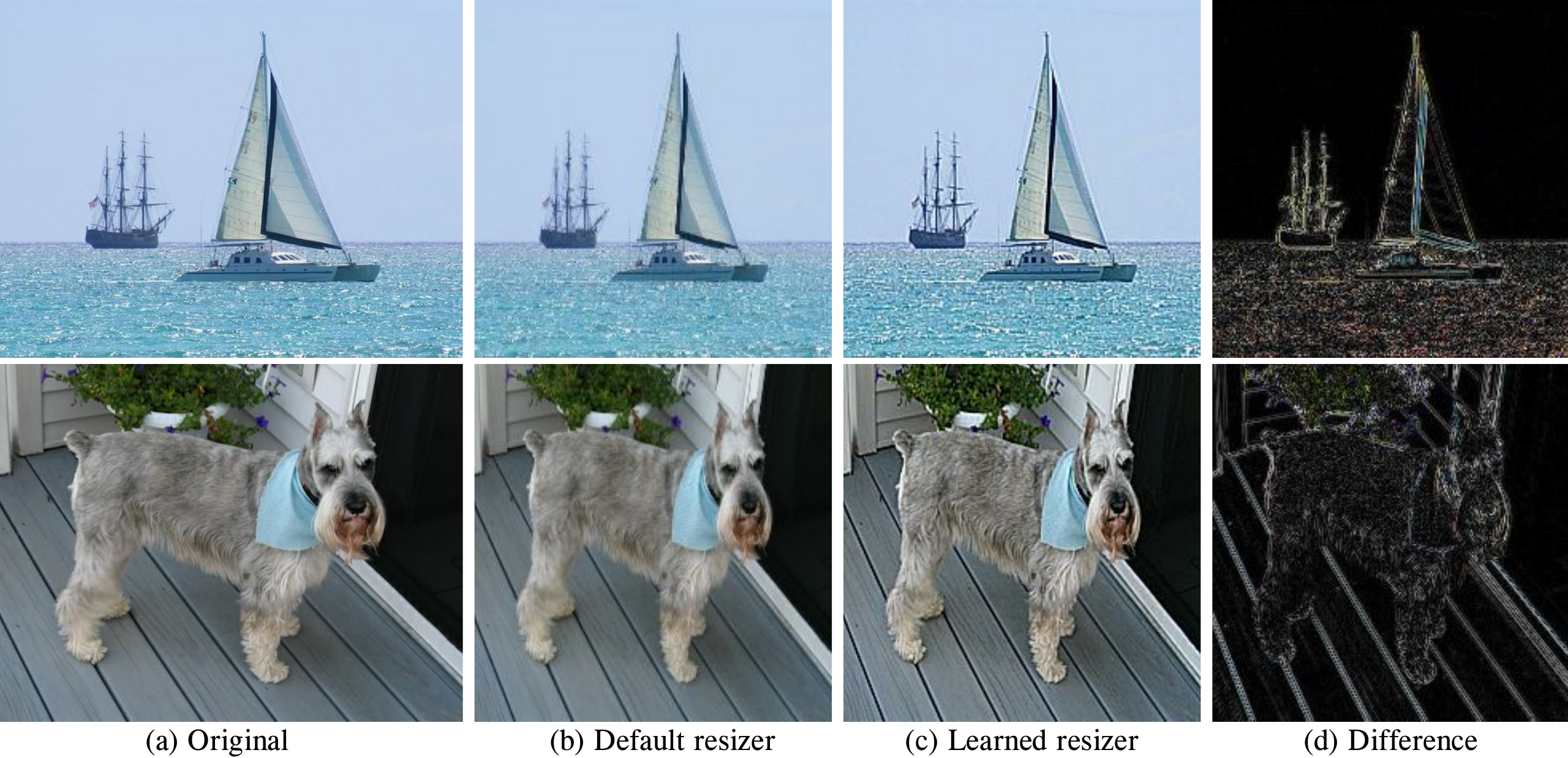}
    %  \caption{Accuracy \vs FLOPs performance scaling curve under ImageNet-1K training setting at various input resolution.}
    %  \label{fig:imagenet-flops}
 \end{subfigure}
 \vspace{-2mm}
\caption{Visualizations of the learned MULLER resizer for ResNet-50. Here the default resizer is an (anti-aliased) \texttt{AREA} resizer in Tensorflow. (d) shows the difference of  the learned and the default resizers.}
\label{fig:muller-vis}
\vspace{-2mm}
\end{figure*}

\noindent\textbf{Image Quality Assessment.}
We base our experiment on the AVA dataset~\cite{murray2012ava}, which includes ~250K images rated by amateur photographers. Each image in the dataset is associated with a histogram of ratings from an average of 200 raters. Image quality and aesthetic assessment is a task that is sensitive to downscaling~\cite{ke2021musiq}, as downscaling may negatively impact visual quality attributes such as sharpness. We use the Earth Mover’s Distance (EMD) as our training loss, similar to previous work of~\cite{talebi2018nima}.

Our results are shown in~\cref{tab:iqa-comparison}. We report the Pearson linear correlation coefficient (PLCC) of the predicted and ground truth mean ratings as our evaluation metric. As can be seen, the proposed resizer improves the performance of MaxViT beyond the existing methods such as MUSIQ~\cite{ke2021musiq}. Note that in contrast to MUSIQ which uses multi-scale input augmentations, MULLER+MaxViT only requires a single low-resolution input from the resizer.

\begin{table}[!t]
\centering
% \footnotesize
% \setlength{\tabcolsep}{4pt}
\renewcommand{\arraystretch}{1}
\begin{tabular}{ccc|cc}
% \hline
% \rowcolor[gray]{0.95}
\multirow{1}{*}{nlayers}  & \multirow{1}{*}{ksize} &
\multirow{1}{*}{std} &
\multirow{1}{*}{Top-1 Acc} & FLOPs \\
% &&&&\\
% \rowcolor[gray]{0.95}
\toprule
% 2 & 5 & 1.0 & 86.37 & 24.23B \\
% 3 & 5 & 1.0 & 86.41 & 24.24B \\
% 4 & 5 & 1.0 & 86.30 & 24.25B \\
% 6 & 5 & 1.0 & 86.43 & 24.27B \\
% \hline
% 2 & 3 & 1.0 & 86.37 & 24.23B \\
% 2 & 7 & 1.0 & 86.36 & 24.24B \\
% \hline
% 2 & 5 & 1.5 & 86.34 & 24.24B\\
% 2 & 5 & 2.0 & 86.41 & 24.24B \\
2 & 5 & 1.0 & 85.58 & 24.23B \\
3 & 5 & 1.0 & 85.52 & 24.24B \\
4 & 5 & 1.0 & 85.50 & 24.25B \\
6 & 5 & 1.0 & 85.59 & 24.27B \\
\hline
2 & 3 & 1.0 & 85.48 & 24.23B \\
2 & 7 & 1.0 & 85.57 & 24.24B \\
\hline
2 & 5 & 1.5 & 85.56 & 24.24B\\
2 & 5 & 2.0 & 85.53 & 24.24B \\
% \hline
% \bottomrule
\end{tabular}
\caption{\textbf{Hyperparameter sweep for MULLER.}}
\label{tab:ablation-hyperparams}
\end{table}
\begin{table}[!t]
\centering
% \footnotesize
% \setlength{\tabcolsep}{4pt}
\renewcommand{\arraystretch}{1}

\begin{tabular}{cc|cc}
% \hline
% \rowcolor[gray]{0.95}
\multirow{1}{*}{input size}  & \multirow{1}{*}{output size} &
\multirow{1}{*}{Top-1 Acc} & FLOPs \\
% &&&&\\
% \rowcolor[gray]{0.95}
\toprule
384 & 224 & 85.45 & 24.22B \\
512 & 224 & 85.58 & 24.23B \\
768 & 224 & 85.56 & 24.26B \\
\hline
512 & 384 & 86.42 & 74.25B \\
768 & 384 & 86.44 & 74.28B \\
\hline
768 & 512 & 86.68 & 138.57B\\
1024 & 512 & 86.67 & 138.60B\\
% \bottomrule
\end{tabular}
\caption{\textbf{Effects of input and output size for MULLER} using MaxViT-Base as test backkbone. Note that the ouput size is the image size seen by the backbone.}
\label{tab:ablation-input-output-size}
\end{table}

\subsection{Ablation}

\noindent\textbf{Hyperparameters.}
There are three hyperparameters in the design of MULLER: $\{k, hsize , stddev\}$, which denote the number of layers, the kernel size of the Gaussian filters $\{\mathbf{W}_1,\mathbf{W}_2,...,\mathbf{W}_k\}$, and the standard deviation of the Gaussian filters.
To understand the effect of these hyperparameters, we conduct an ablation study.
As shown in \cref{tab:ablation-hyperparams}, we found that MULLER is quite insensitive to the selection of parameters, and thus we recommend to use a simple set of parameters to save computation.

\noindent\textbf{Effects of image size.}
It is known that image size can significantly affect the recognition performance.
We evaluate the effect of varying input and output sizes of MULLER using MaxViT-B in \cref{tab:ablation-input-output-size}.
Note that the output size of MULLER is indeed the size seen by the backbone, so higher output size typically corresponds to improved accuracy.
We observe that using higher input resolutions (\eg, 3$\times$) for MULLER does not yield any further performance gain beyond the baseline setting.
Nonetheless, we find that adopting a reasonably large resolution (\eg 1.6$\sim$2.5$\times$) is necessary to achieve the expected performance. It is worth highlighting that this is perhaps impacted by the original resolution of images in the benchmark dataset.

\subsection{Visualization}

We visualize the behavior of the learned resizer in \cref{fig:muller-vis}.
As can be seen, the MULLER resizer is learned to boost details or textures of the images, while also enhancing the image contrast.
These effects can preserve more visual information in the downscaled images over naive resizing, thus making the classification model learn better.
As compared to the previous less-controlled resizer~\cite{talebi2021learning}, MULLER achieves a better balance of human and machine perceptual qualities, due to the strong regularization imbued in its Laplacian-inspired design.
We also point out that training MULLER with aliased inputs may produce relatively less sharper images in comparison to \cref{fig:muller-vis}. We refer the reader to \cref{sec:supp-vis} for visual examples.
\section{Concluding Remarks}

In this paper, we introduce MULLER, an extremely simple and light learned resizer, using multilayer Laplacian decomposition.
The proposed resizer only contains 4 trainable parameters with negligible training and inference costs. This allows deploying the resizer as a thumbnail generator to produce optimally downscaled images for sending to remote inference servers, or alternatively as a server side resizer that reduces the inference cost without the necessity of changing the backbone architecture.
We show that MULLER not only pushes forward the limit of the state-of-the-art vision Transformer MaxViT on ImageNet classification, but it also consistently improves across a range of widely used architectures, including EfficientNet, MobileNet, and ResNet.
Additionally, we provide experiments to substantiate the efficacy of MULLER for various downstream tasks, such as object detection, segmentation, and image quality assessment.
As compared to previous methods, MULLER enjoys remarkable generalization ability, owing to the strong regularization provided by its multilayer bandpass design.
We believe that our work will inspire future research in this critically important direction: how to better preprocess images for vision tasks.

\noindent\textbf{Limitations and future works.}
We note that if the higher-resolution inputs fail to boost the performance of an specific task, we cannot reasonably expect the learned resizer to provide a substantial performance boost either.
Another potential future direction is to train a universal learned resizer that can be a drop-in replacement of the off-the-shelf resizers in existing machine learning frameworks, without necessitating the joint re-training of the backbones.

{\small
\bibliographystyle{ieee_fullname}
\bibliography{egbib}
}
\clearpage % This will ensure the Appendix starts on a new page

\onecolumn % Switch to one-column layout

\appendix
\section*{Appendix}
This appendix is organized as follows:

\begin{itemize}
    \item We present detailed experimental settings and hyperparameters for image classification, object detection and segmentation, and image quality experiments in \cref{sec:supp-detailed-configs}.
    \item Additional experimental results of MULLER resizer with respect to comparisons with previous works and the generalization are provided in ~\cref{sec:supp-additional-results}.
    \item The discussion of the anti-aliasing effect as well as a comprehensive visualization are given in \cref{sec:supp-anti-alias} and \cref{sec:supp-vis}.
\end{itemize}

\section{Experimental Settings}
\label{sec:supp-detailed-configs}

\subsection{ImageNet Classification}
\label{ssec:imagenet-settings}
We provide the experimental settings for both pre-training and fine-tuning MaxViT models on ImageNet-1K, detailed in~\cref{tab:supp-i1k-exp-settings}.
All the MaxViT variants employed similar hyperparameters except for that the the stochastic depth rate was tuned for each setting.
It should be noted that we first pre-trained the backbone on ImageNet-1k/-21k/JFT with 300/90/14 epochs at a resolution of $224\times 224$.
Subsequently, the backbone was jointly fine-tuned with MULLER plugged-in at a higher resolution for an additional 30 epochs.

\begin{table}[!ht]
% \scriptsize
\centering
\setlength{\tabcolsep}{1pt}
\renewcommand{\arraystretch}{1.}

\begin{tabular}{l|cc|cc|cc}
% \toprule
\multirow{3}{*}{Hyperparameter}  & \multicolumn{2}{c|}{\textbf{ImageNet-1K}} & \multicolumn{2}{c|}{\textbf{ImageNet-21K}} &
\multicolumn{2}{c}{\textbf{JFT-300M}} \\
& Pre-train & Fine-tune(+MULR) & Pre-train & Fine-tune(+MULR) & Pre-train & Fine-tune(+MULR) \\
& \multicolumn{2}{c|}{(MaxViT-T/S/B/L)}
& \multicolumn{2}{c|}{(MaxViT-B/L/XL)}
& \multicolumn{2}{c}{(MaxViT-B/L/XL)} \\
\toprule
Stochastic depth & \multicolumn{1}{c}{$0.2/0.3/0.4/0.6$} &
\multicolumn{1}{c|}{$0.3/0.5/0.7/0.95$} & \multicolumn{1}{c}{$0.3/0.4/0.6$} &
\multicolumn{1}{c|}{$0.4/0.5/0.9$} &
\multicolumn{1}{c}{$0.0/0.0/0.0$} & \multicolumn{1}{c}{$0.1/0.2/0.1$} \\
Center crop  & True & False & True & False & True & False \\
RandAugment & 2, 15 & 2, 15 & 2, 5 & 2, 15 & 2, 5 & 2, 15 \\
Mixup alpha & 0.8 & 0.8 & None & None & None & None  \\
Loss type & Softmax & Softmax & Sigmoid & Softmax & Sigmoid & Softmax \\
Label smoothing & 0.1 & 0.1 & 0.0001 & 0.1 & 0 & 0.1 \\
Train epochs & 300 & 30 & 90 & 30 & 14 & 30 \\
Train batch size & 4096 & 512 & 4096 & 512 & 4096 & 512 \\
Optimizer type & AdamW & AdamW & AdamW & AdamW & AdamW & AdamW \\
Peak learning rate & 3e-3 & 5e-5 & 1e-3 & 5e-5 & 1e-3 & 5e-5 \\
Min learning rate & 1e-5 & 5e-5 & 1e-5 & 5e-5 & 1e-5 & 5e-5 \\
Warm-up & 10K steps & None & 5 epochs & None & 20K steps & None \\
LR decay schedule & Cosine & None & Linear & None & Linear & None \\
Weight decay rate & 0.05 & 1e-8 & 0.01 & 1e-8 & 0.01 & 1e-8 \\
Gradient clip & 1.0 & 1.0 & 1.0 & 1.0 & 1.0 & 1.0 \\
EMA decay rate & None & 0.9999 & None & 0.9999 & None & 0.9999 \\
\bottomrule
\end{tabular}
\caption{\textbf{Detailed hyperparameters used in ImageNet-1K experiments.} Multiple values separated by `$/$' are for each model size respectively.}
\label{tab:supp-i1k-exp-settings}
\end{table}

\subsection{Object Detection and Segmentation}
\label{ssec:coco-settings}

We evaluated MaxViT on the COCO2017~\cite{lin2014microsoft} object bounding box detection and instance segmentation task.
The dataset comprises 118K training and 5K validation samples.
All MaxViT backbones were pretrained on the ImageNet-1k dataset at a resolution of $224\times224$ following the same training protocol detailed in~\cref{ssec:imagenet-settings}.
These pretrained checkpoints were then used as the warm-up weights for fine-tuning on the detection and segmentation tasks.
Note that for both tasks, the input images were resized to $896\times896$ before being fed into the MULLER resizer.
The backbone was actually receiving a $640\times640$ resolution images for generating the box proposals.
The training was conducted with a batch size of 256, using the AdamW~\cite{loshchilov2017decoupled} optimizer with learning rate of 3e-3, and stochastic depth of $0.3, 0.5, 0.8$ for MaxViT-T/S/B backbones, respectively.

\subsection{Image Quality Assessment}
\label{ssec:image-quality-settings}

We trained and evaluated the MaxViT model on the AVA benchmark~\cite{murray2012ava}. Similar to \cite{talebi2018nima,ke2021musiq}.
We pre-train MaxViT for resolutions: $224 \times 224$.
Then we initialized the model with ImageNet-1K $224\times224$ pre-trained weights and fine-tune it with MULLER resizer. The weight and bias momentums are set to 0.9, and a dropout rate of 0.75 is applied on the
last layer of the baseline network. We use an initial learning rate of 1e-3, exponentially decayed with decay factor 0.9 every 10 epochs. We set the stochastic depth rate to 0.5.

\section{Additional Experimental Results}
\label{sec:supp-additional-results}

\subsection{Comparisons to Previous Resizer}

We compare the proposed MULLER resizer against the previous learned resizer with residual convolution blocks~\cite{talebi2021learning}.
As shown in~\cref{tab:supp-comparison-to-residual}, fine-tuning with MULLER performs as effective as, and sometimes better than the previous heavier residual resizer.
Furthermore, we note that MULLER is two orders-of-magnitude cheaper in inference cost (FLOPs), which further saves up to 52\% training cost on TPUs, depending on the model size.
Thus, MULLER is a promising `green' machine learning model that can be easily integrated into various applications without incurring additional costs.
Another benefit of MULLER over~\cite{talebi2021learning} is that MULLER is restricted to generating images that are more comprehensible to humans, despite being trained only for machine vision.
This may be attributed to the bandpass design of the multilayer Laplacian filters employed in MULLER.

\begin{table}[!ht]
\centering
% \footnotesize
\setlength{\tabcolsep}{4pt}
\renewcommand{\arraystretch}{1}

\begin{tabular}{l|ccccc}
% \hline
% \rowcolor[gray]{0.95}
\multirow{2}{*}{Model} & 
\multirow{2}{*}{
\begin{tabular}{c}Size\end{tabular}
} & 
\multirow{2}{*}{
\begin{tabular}{c}Infer cost\\(FLOPs)\end{tabular}
} & 
\multirow{2}{*}{
\begin{tabular}{c}Train cost\\(TPUv3 hrs)\end{tabular}
} &
\multirow{2}{*}{
\begin{tabular}{c}top-1\\ accuracy \end{tabular}
} \\
&&&&\\
% \rowcolor[gray]{0.95}
\toprule

MaxViT-T & 224 & 5.6B & - & 83.62  \\
+Residual~\cite{talebi2021learning}$_{512\rightarrow 224}$ & 224 & 6.8B & 2.8 & 83.93  \\
\rowcolor[gray]{0.95}+MULLER$_{512\rightarrow 224}$ & 224 & 5.6B & 1.9 & 83.95  \\ \hline

MaxViT-S & 224 & 11.7B & - &  84.45 \\
+Residual~\cite{talebi2021learning}$_{512\rightarrow 224}$ & 224 & 12.9B & 4.2 & 84.95  \\
\rowcolor[gray]{0.95}+MULLER$_{512\rightarrow 224}$ & 224 & 11.7B & 2 & 85.95  \\
\hline

MaxViT-B & 224 & 23.4B & - & 84.95 \\
+Residual~\cite{talebi2021learning}$_{512\rightarrow 224}$ & 224 & 25.4B & 6 & 85.48  \\
\rowcolor[gray]{0.95}+MULLER$_{512\rightarrow 224}$ & 224 & 23.4B & 3.5 & 85.58  \\

MaxViT-L & 224 & 43.9B & - & 85.17 \\
+Residual~\cite{talebi2021learning}$_{512\rightarrow 224}$ & 224 & 45.1B & 7.7 & 85.73  \\
\rowcolor[gray]{0.95}+MULLER$_{512\rightarrow 224}$ & 224 & 43.9B  & 5.0 & 85.68 \\

% \hline
\bottomrule
\end{tabular}
\caption{\normalsize \textbf{Performance comparison against previous residual resizer~\cite{talebi2021learning}.} }
\label{tab:supp-comparison-to-residual}
\end{table}

\subsection{Transferability Experiments.}

We also examine the generalization ability of the learned resizer across different MaxViT model variants.
Specifically, we take the learned resizer parameters from one MaxViT variant, and directly test it on another variant.
As can be seen in \cref{tab:supp-imagenet1k-cross-validation-maxvit}, the learned resizer generalizes very well across different MaxViT model scales.
% which is consistent to the fact in \cref{tab:imagenet1k-backbones-params} that learned weights are closer for MaxViT-T and -S, and for MaxViT-B and -L, respectively.
The average top-1 accuracy drop is less than $0.06$ when using different learned weights, indicating great transferrability of the MULLER resizer.

\begin{table}[!ht]
\centering
% \footnotesize
% \setlength{\tabcolsep}{1pt}
\renewcommand{\arraystretch}{1}

\begin{tabular}{l|cccc}
% \hline
% \rowcolor[gray]{0.95}
\multirow{1}{*}{Model} &
MaxViT-T & 
MaxViT-S & 
MaxViT-B &
MaxViT-L 
\\ 
% \rowcolor[gray]{0.95}
\toprule

MULLER$_\text{M-T}$ & 83.95 & 84.91 & 85.61 & 85.68 \\
MULLER$_\text{M-S}$ & 83.96 & 84.95 & 85.61 & 85.68 \\
MULLER$_\text{M-B}$ & 83.97 & 84.89 & 85.58 & 85.69 \\
MULLER$_\text{M-L}$ & 83.95 & 84.91 & 85.61 & 85.68 \\
% \hline
\bottomrule
\end{tabular}
\caption{\normalsize \textbf{Cross-model validation of the MULLER resizer for ImageNet-1K on MaxViT variants.} These values represent the top-1 accuracy of a given backbone tested with various MULLER resizers.}
\label{tab:supp-imagenet1k-cross-validation-maxvit}
\end{table}

\subsection{The Effect of Base Resize Method.}
We conduct another ablation study to inspect the effects of the base resize method used inside MULLER.
It is worth highlighting that this is the resizer used in MULLER, and it may or may not be different than the `default resizier' mentioned in the main paper.
Since we run all the experiments on TPU devices, we have found that only \texttt{bilinear} and \texttt{nearest} resizers are compilable.
As demonstrated in~\cref{tab:supp-base-resize}, using nearest method as the base resizer yields similar performance as compared to the default bilinear method.
We thus hypothesize that the choice of the base resize method used in MULLER does not significantly affect the performance of the model.

\begin{table}[!ht]
\centering
% \footnotesize
% \setlength{\tabcolsep}{1pt}
\renewcommand{\arraystretch}{1}

\begin{tabular}{lcccc}
% \hline
% \rowcolor[gray]{0.95}
\multirow{1}{*}{Model} & \multirow{1}{*}{Resize method} &
\multirow{1}{*}{TPU compilable?} &
\multirow{1}{*}{Top-1 acc.} \\
% &&&&\\
\toprule
MaxViT-B & Bilinear & Yes & 85.58 \\
& Nearest & Yes & 85.54 \\ 
% & Area & No & - \\
& Bicubic & No & -\\
& Lanczos & No & -\\
\bottomrule
\end{tabular}
\caption{\normalsize \textbf{Effects of base resize method used in MULLER.}}
\label{tab:supp-base-resize}
\end{table}

% Bicubic and lanczos are not compilation on TPU.

\section{On Anti-Aliasing}
\label{sec:supp-anti-alias}
We now investigate the effects of anti-aliasing on the input images to MULLER resizer.
Our experiments reveal that while removing anti-aliasing does not affect the overall performance gain obtained by MULLER, the learned parameters may differ.
As shown in~\cref{tab:supp-antialiasing}, the learned parameters for each backbone have a slight shift in the weights and biases.
However, these place no effects on the fine-tuning performances.

\begin{figure*}[!tb]
 \centering
 \begin{subfigure}[b]{0.99\textwidth}
     \centering
     \includegraphics[width=0.99\textwidth]{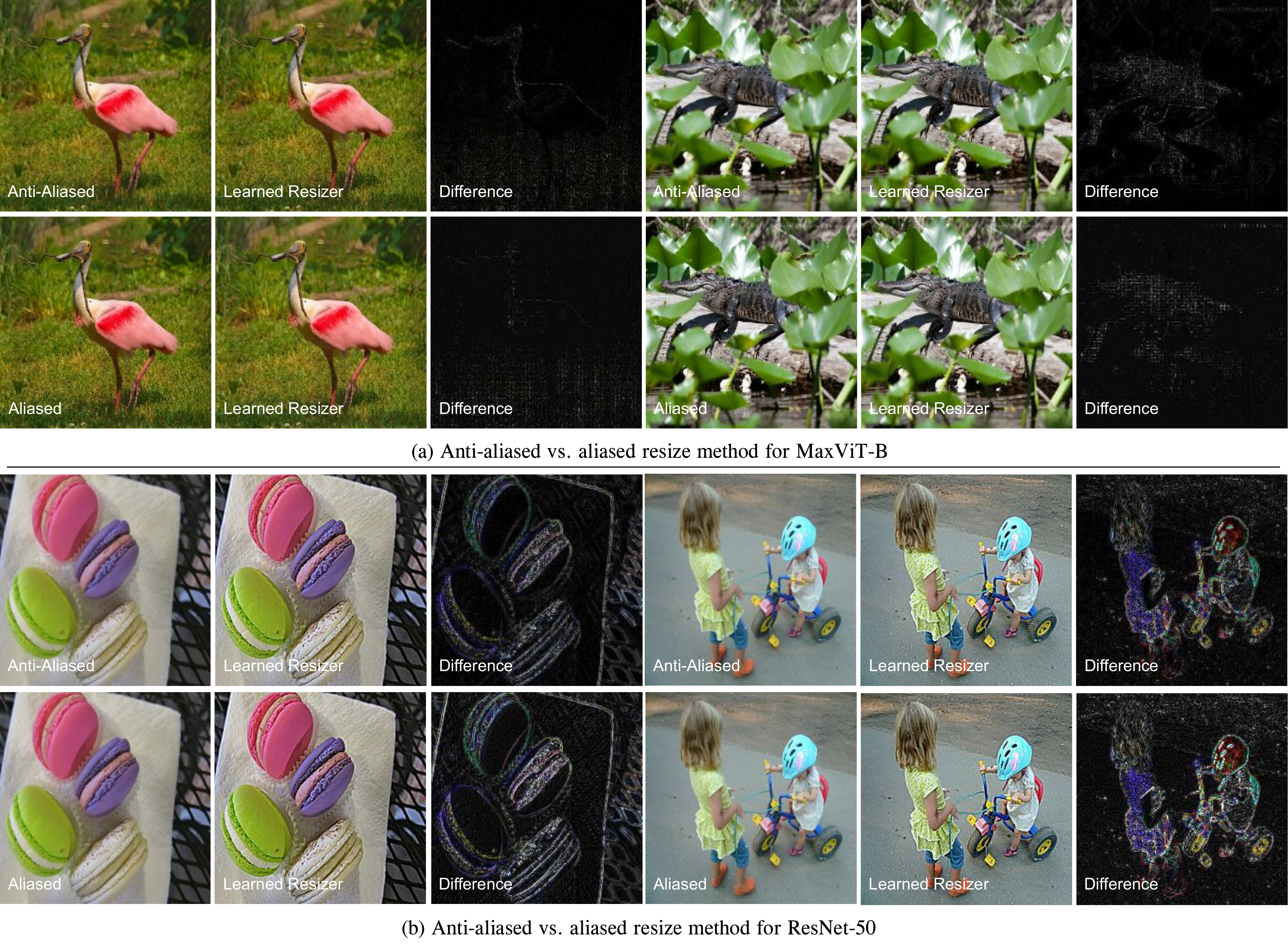}
    %  \caption{Accuracy \vs FLOPs performance scaling curve under ImageNet-1K training setting at various input resolution.}
    %  \label{fig:imagenet-flops}
 \end{subfigure}
 \vspace{-2mm}
\caption{Visualization of the impact of anti-aliasing for the input image of MULLER. (a) shows examples for MaxViT-B, while (b) demonstrates those for ResNet-50.}
\label{fig:supp-muller-anti-alias}
\vspace{-2mm}
\end{figure*}

We do observe that anti-aliasing may impact the behavior of the learned resizer in terms of visualizations.
For instance, as shown in~\cref{fig:supp-muller-anti-alias}, (a) when anti-aliasing is enabled for MaxViT-B, MULLER learns to enhance the contrast/details of the image to some extent; if the input image is aliased, nevertheless, MULLER learns to reduce the `aliased effects'. In other words, the difference image displays some patterns similar to the aliasing effects in the resized image.
(b) as for ResNet-50, it may be seen that MULLER learns to boost details even more for aliased inputs than the anti-aliased.
Both effects have not been observed to significantly impact the performances, though.

\begin{table}[!ht]
\centering
% \footnotesize
\setlength{\tabcolsep}{4pt}
\renewcommand{\arraystretch}{1}

\begin{tabular}{c|l|cccccc}
% \hline
% \rowcolor[gray]{0.95}
% \multirow{1}{*}{Resize method} &
\multirow{1}{*}{Anti-aliasing?} &
\multirow{1}{*}{Model} & \multirow{1}{*}{\begin{tabular}{@{}c@{}}$\alpha_1$\end{tabular}} & \multirow{1}{*}{$\beta_1$} &
\multirow{1}{*}{\begin{tabular}{@{}c@{}}$\alpha_2$\\ \end{tabular}} &
\multirow{1}{*}{$\beta_2$} &
top-1 acc.
\\
% &&&&\\
% \rowcolor[gray]{0.95}
\toprule
% MaxViT-T~\cite{tu2022maxvit} & 0.310 & 0.021 & -0.738 & 0.020 \\
% MaxViT-S~\cite{tu2022maxvit} & 0.359 & -0.003 & -0.779 & -0.003 \\
% MaxViT-B~\cite{tu2022maxvit} & 1.022 & -0.037 & -1.185 & -0.026  \\
% MaxViT-L~\cite{tu2022maxvit} & 1.000 & -0.015 & -1.122 & -0.006 \\
% \hline
% \multirow{3}{*}{Area} &
\multirow{3}{*}{Yes} &
EffNet-B0~\cite{tan2019efficientnet} & 1.715 & 0.088 & -8.41 & 0.001 & 78.2 \\
&   MobileNet-v2~\cite{sandler2018mobilenetv2} & 1.480 & 0.174 & -5.25 & -0.058 & 71.8 \\
&  ResNet-50~\cite{he2016deep} & 1.892 & -0.014 & -11.295 & 0.003 & 76.2 \\
\midrule
% \multirow{3}{*}{Bilinear} &
\multirow{3}{*}{No} & EffNet-B0~\cite{tan2019efficientnet} & 1.632 & -0.014 & -7.265 & 0.026 & 78.2 \\
&  MobileNet-v2~\cite{sandler2018mobilenetv2} & 1.792 & 0.269 & -7.514 & -0.077 & 71.7 \\
& ResNet-50~\cite{he2016deep} & 1.687 & -0.039 & -12.637 & 0.015 & 76.2 \\

% \hline
\bottomrule
\end{tabular}
\caption{\normalsize {The learned MULLER parameters for different backbone models train on ImageNet-1k. Top 3 rows show results using anti-aliased resizer while bottom 3 rows depict aliased resizing.}.}
\label{tab:supp-antialiasing}
\end{table}

\section{Visualization}
\label{sec:supp-vis}

\begin{figure*}[!tb]
 \centering
 \begin{subfigure}[b]{0.99\textwidth}
     \centering
     \includegraphics[width=0.99\textwidth]{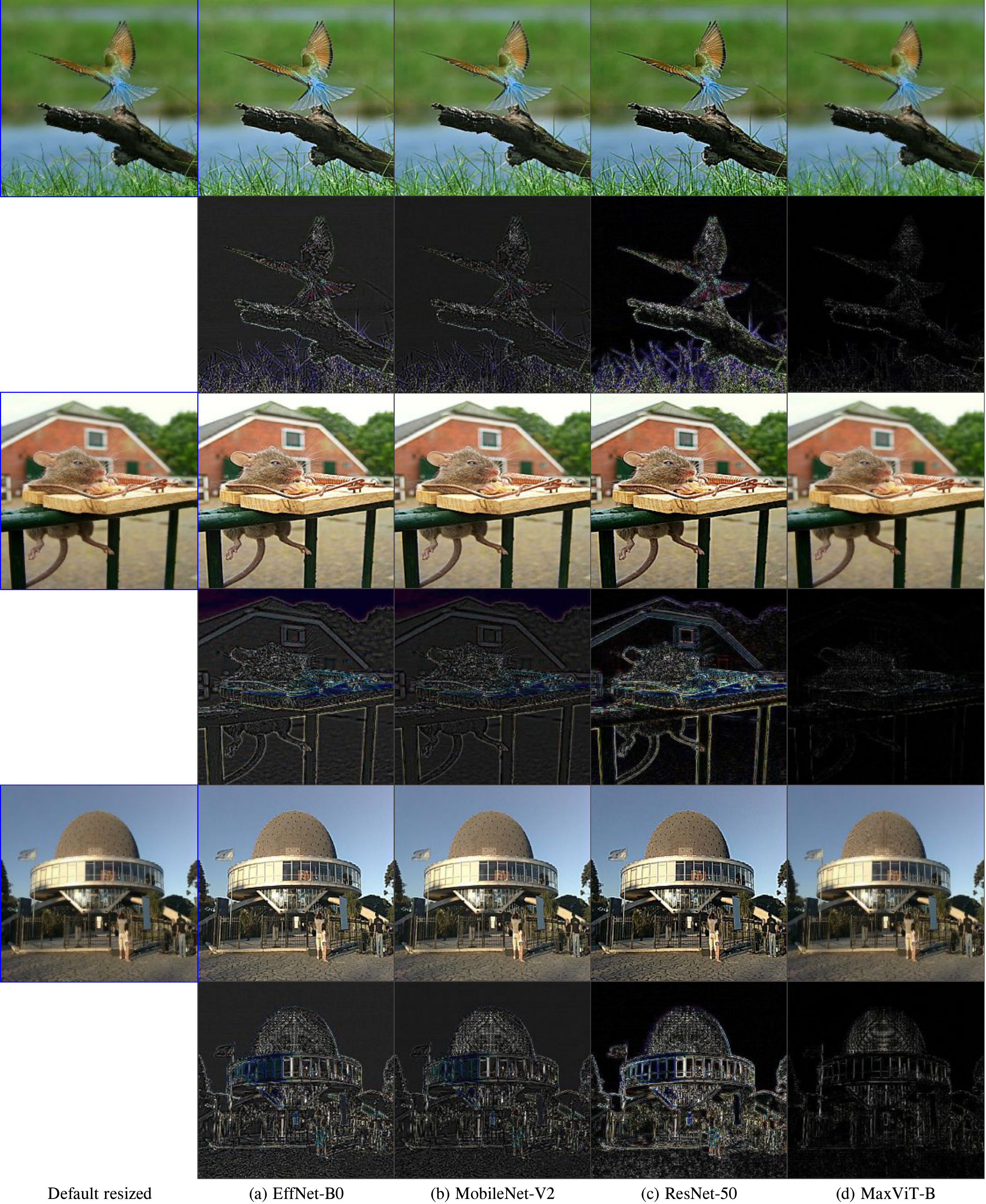}
    %  \caption{Accuracy \vs FLOPs performance scaling curve under ImageNet-1K training setting at various input resolution.}
    %  \label{fig:imagenet-flops}
 \end{subfigure}
 \vspace{-2mm}
\caption{Visualizations of the MULLER resizer for (a) EffNet-B0, (b) MobileNet-V2, (c) ResNet-50, and (d) MaxViT-B. Here the default resizer is an anti-aliased resizer. Below each resized image shows the difference with the default resizer.}
\label{fig:supp-muller-vis-1}
\vspace{-2mm}
\end{figure*}

\begin{figure*}[!tb]
 \centering
 \begin{subfigure}[b]{0.99\textwidth}
     \centering
     \includegraphics[width=0.99\textwidth]{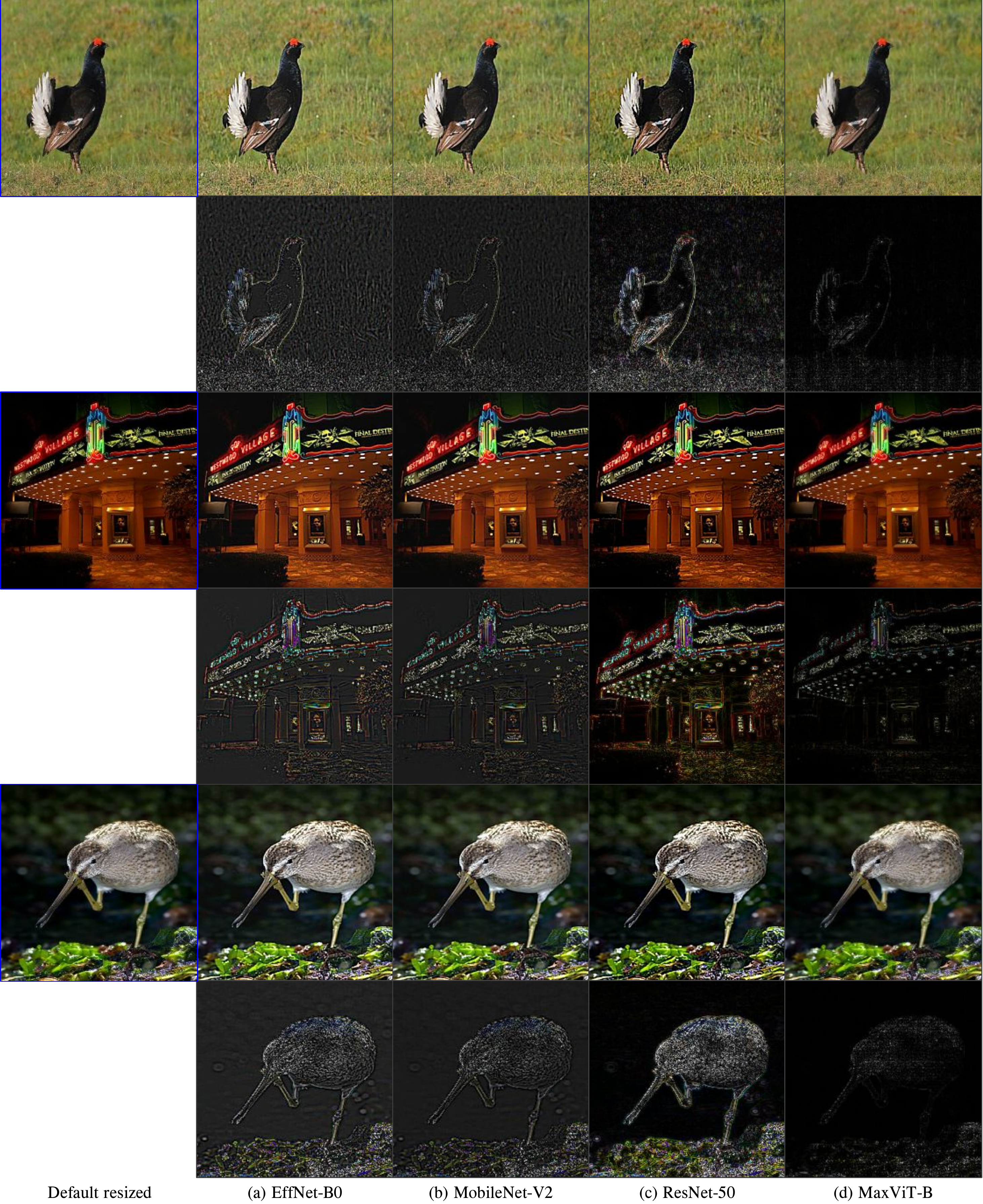}
    %  \caption{Accuracy \vs FLOPs performance scaling curve under ImageNet-1K training setting at various input resolution.}
    %  \label{fig:imagenet-flops}
 \end{subfigure}
 \vspace{-2mm}
\caption{Visualizations of the MULLER resizer for (a) EffNet-B0, (b) MobileNet-V2, (c) ResNet-50, and (d) MaxViT-B. Here the default resizer is an anti-aliased resizer. Below each resized image shows the difference with the default resizer.}
\label{fig:supp-muller-vis-2}
\vspace{-2mm}
\end{figure*}

\cref{fig:supp-muller-vis-1,fig:supp-muller-vis-2} illustrate some additional visualization results of the learned MULLER resizer for various backbones, including (a) EffNet-B0, (b) MobileNet-V2, (c) ResNet-50, and (d) MaxViT-B, arranged in ascending order of model complexity.
A few of observations can be made:
(1) On all the models, the MULLER resizer learns to boost the details/contrast of the image, albeit with varying degrees;
(2) As evident from the performance gain of the vision models, the embedded information in the MULLER resized images is machine-friendly, and contributes to a more effective learning of the backbone;
(3) Due to the highly regularized design of the resizer, the outputs of MULLER remain highly perceivable by human (in some cases even look perceptually superior), even though MULLER is purely trained for machine vision.

\end{document}